\documentclass[runningheads]{llncs}
\usepackage[T1]{fontenc}
\usepackage{graphicx}
\usepackage{algorithm}
\usepackage[noend]{algpseudocode}
\usepackage{booktabs}
\usepackage{multirow}

\newif\iflncs
\lncsfalse   % Full version

\usepackage{xspace}
\usepackage{comment}

\usepackage{subcaption}
\usepackage{amsmath}
\usepackage{mathtools}
\usepackage{amssymb,amsfonts}
\usepackage[hypertexnames=false,setpagesize=false]{hyperref}
\usepackage[nameinlink,capitalize,noabbrev]{cleveref}
\usepackage{autonum}
\usepackage{bm}
\usepackage{color}
\usepackage{tabularx}

\crefname{equation}{Eq.}{Eqs.}
\crefname{figure}{Fig.}{Figs.}

\newcommand{\R}{\mathbb{R}}

\newcommand{\vp}[1]{^{(#1)}}

\newcommand{\intersectionpoint}{intersection point\xspace}
\newcommand{\intersectionspace}{intersection space\xspace}
\newcommand{\intersectionpoints}{intersection points\xspace}
\newcommand{\intersectionspaces}{intersection spaces\xspace}

\newcommand{\bmb}{\bm{b}}

\newcommand{\bme}{\bm{e}}

\newcommand{\bmn}{\bm{n}}

\newcommand{\bmp}{\bm{p}}
\newcommand{\bmq}{\bm{q}}

\newcommand{\bms}{\bm{s}}

\newcommand{\bmv}{\bm{v}}
\newcommand{\bmw}{\bm{w}}
\newcommand{\bmx}{\bm{x}}
\newcommand{\bmy}{\bm{y}}
\newcommand{\bmz}{\bm{z}}

\newcommand{\bfD}{\mathbf{D}}

\newcommand{\bfF}{\mathbf{F}}
\newcommand{\bfG}{\mathbf{G}}

\newcommand{\bfI}{\mathbf{I}}

\newcommand{\bfN}{\mathbf{N}}

\newcommand{\bfW}{\mathbf{W}}

\renewcommand{\ker}{\mathrm{Ker}}

\newcommand{\Span}{\mathrm{span}}

\begin{document}
%
% \title{Polynomial Time Cryptanalytic Extraction in EC25 is Exponentially Hard for Real DNNs}
\title{Is the Hard-Label Cryptanalytic Model Extraction Really Polynomial?}
% \author{Akira Ito\orcidID{0000-0002-4602-7570} \and
% Takayuki Miura\orcidID{0000-0001-8694-312X} \and
% Yosuke Todo\orcidID{0000-0002-6839-4777}}
 \author{Akira Ito\inst{1} \and
 Takayuki Miura\inst{2} \and
 Yosuke Todo\inst{2}}
 \authorrunning{A. Ito et al.}
 \institute{
Tohoku University \\ 
  2--1--1 Katahira, Aoba-ku, Sendai-shi, 980-8577, Japan \\
 \email{akira.ito.b1@tohoku.ac.jp}, 
 \and
 NTT Social Informatics Laboratories, \\ 
 3--9--11 Midori-cho,
  Musashino-shi, Tokyo, 180-8585, Japan \\
 \email{tkyk.miura@ntt.com},
 \email{yosuke.todo@ntt.com}
 }
\maketitle              % typeset the header of the contribution
\begin{abstract}
Deep Neural Networks (DNNs) have attracted significant attention, and their internal models are regarded as valuable intellectual assets.
Extracting a model via oracle access to a DNN is conceptually similar to extracting a secret key via oracle access to a block cipher. 
Consequently, cryptanalytic techniques, particularly differential-like attacks, have been actively explored recently. 
ReLU-based DNNs are the most common architectures. 
While early works (e.g., Crypto 2020, Eurocrypt 2024) assume access to exact output logits, more recent works (e.g., Asiacrypt 2024, Eurocrypt 2025) focus on the hard-label setting, where the attacker observes only the final classification result (e.g., ``dog'' or ``car''). 
Notably, Carlini et al. (Eurocrypt 2025) demonstrated that model extraction is feasible in polynomial time even under this restricted setting.

In this paper, we show that a key assumption of their attack becomes increasingly unrealistic as the target depth grows. While prior works have noted neurons whose activation states rarely change, we analyze their concrete impact on hard-label extraction: even a single neuron that is (almost) always active can prevent the attack from proceeding unless its parameters are recovered, and ignoring it inevitably incurs a non-negligible error. 
A straightforward solution is to recover these parameters from a state switch, but observing such a switch becomes exponentially harder as the depth increases, implying that hard-label extraction is not always polynomial time.
To address this limitation, we propose \emph{cross-layer extraction}. Rather than extracting the secret parameters (e.g., weights and biases) directly, we exploit cross-layer interactions to recover them from deeper layers, reducing query complexity and addressing limitations of existing model extraction approaches. 
\iflncs
\else
The source code is available at \url{https://github.com/ECSIS-lab/hard-label-cross-layer-extraction}.
\fi

% We also show that forcing such a neuron to become inactive without incurring this error becomes exponentially harder as the depth increases, implying that hard-label extraction is not always polynomial time.

% In this paper, we first show that the assumptions underlying their attack become increasingly unrealistic as the attack-target depth grows. In practice, satisfying these assumptions requires an exponential number of queries with respect to the attack depth, implying that the attack does not always run in polynomial time.
% To address this critical limitation, we propose a novel attack method called \emph{cross-layer extraction}. Instead of directly extracting the secret parameters (e.g., weights and biases) of a specific neuron, which can incur exponential cost, we exploit cross-layer interactions to extract this information from deeper layers. This technique significantly reduces query complexity and mitigates the limitations of existing model extraction approaches.

\end{abstract}
%
%
%

%%%%%%%%%%%%%%%%%%%%%%%%%%%%%%%%%%%%%%%%%%%%%%%%%%%
%%%%%%%%%%%%%%%%%%%%%%%%%%%%%%%%%%%%%%%%%%%%%%%%%%%

\section{Introduction}
Deep Neural Networks (DNNs) have become one of the most fundamental AI technologies and play an important role in various aspects of our society. Well-trained models can perform complex tasks, such as image classification and natural language processing, often surpassing human-level performance~\cite{DBLP:conf/nips/VaswaniSPUJGKP17,DBLP:conf/ssw/OordDZSVGKSK16,DBLP:conf/icml/RadfordKHRGASAM21}. Therefore, they are regarded as intellectual assets, and their security has become a critical concern. If a (malicious) user is able to extract the learned model, it poses a significant risk to the model provider~\cite{DBLP:conf/icml/MartinelliSGB24,DBLP:conf/uss/JagielskiCBKP20,DBLP:conf/nips/FoersterMSH24,DBLP:conf/iclr/DanielyG23,DBLP:conf/fat/MilliSDH19,DBLP:conf/uss/TramerZJRR16,DBLP:conf/icml/RolnickK20,DBLP:conf/sectl/MiuraSY24}. 

ReLU-based DNNs are widely deployed due to their simplicity and effectiveness. 
Evaluating the feasibility of model extraction, in which an attacker extracts model parameters solely from input-output behavior, is therefore particularly important for ReLU-based DNNs. 
Recently, cryptographers have shown increasing interest in this problem because, from a high-level point of view, ReLU-based DNNs share structural similarities with block ciphers. 
Consequently, cryptanalytic techniques are a promising approach to model extraction. 

\begin{figure}[tb]
    \centering
    \includegraphics[width=\linewidth]{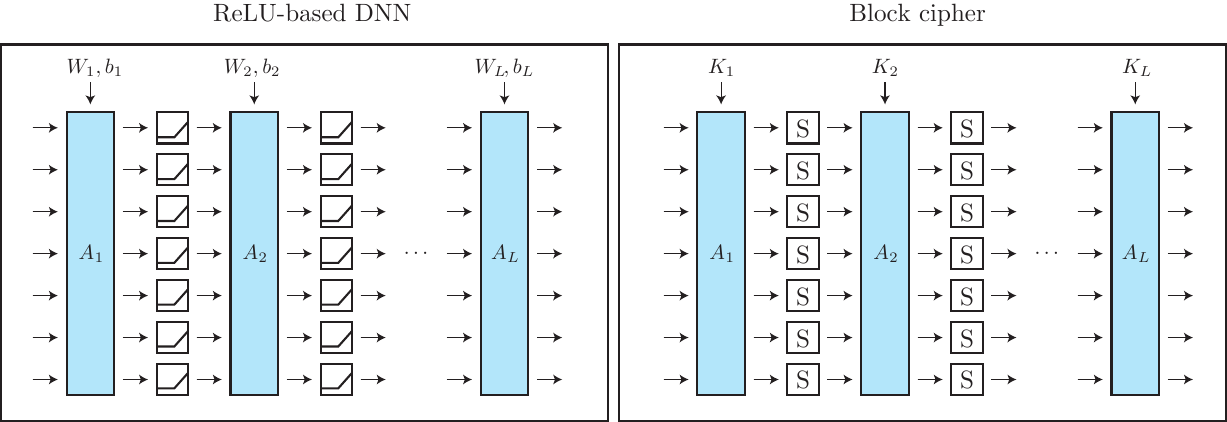}
    \caption{ReLU-based DNN and block cipher}
    \label{fig:relu-bc}
\end{figure}
\cref{fig:relu-bc} illustrates the structural comparison between ReLU-based DNNs and block ciphers. 
As the figure shows, DNNs and block ciphers exhibit notable similarities. 
Both share an iterative structure in which each step consists of a linear (more precisely, affine) layer and a nonlinear layer. In block ciphers, the only nonlinear components are the S-boxes, while in DNNs, the ReLU functions serve as the sole nonlinear elements. 
Common block ciphers such as AES \cite{DBLP:books/sp/DaemenR02} are constructed from public linear layers and S-boxes, with secret subkeys derived from a key schedule. 
More general constructions, such as SASAS \cite{DBLP:journals/joc/BiryukovS10} or ASASA \cite{DBLP:conf/asiacrypt/BiryukovBK14}, incorporate entirely secret affine and nonlinear layers. 
In contrast, the affine layers of DNNs are typically not publicly disclosed, and constitute the core intellectual asset of the model. The multiplication matrices in the linear layers are referred to as \emph{weights}, and the added secret vectors are referred to as \emph{biases}.

On the other hand, it would be overly simplistic to assume that DNNs and block ciphers are identical. When considering model extraction attacks, the most crucial difference lies in the piecewise linearity of DNNs. 
In block ciphers, even a 1-bit (or more generally, an arbitrary) difference significantly alters the nonlinear behavior. To ensure sufficient diffusion, designers guarantee a large number of active S-boxes.
In contrast, let $f$ be a ReLU network. For a given input $\bm x \in \mathbb{R}^{d_0}$, the function $f$ is locally affine around $\bm x$. That is, for sufficiently small $\bm \delta \in \mathbb{R}^{d_0}$, there exists a matrix $\mathbf{F}_{\bm x}$ dependent on $\bm x$, such that  
\[f(\bm x + \bm \delta) = \mathbf F_{\bm x} \bm \delta + f(\bm x). \]
This local linearity is one of the key reasons why polynomial-time model extraction attacks are feasible for DNNs. 

Early works~\cite{DBLP:conf/crypto/CarliniJM20,DBLP:conf/eurocrypt/CanalesMartinezCHRSS24,DBLP:conf/asiacrypt/ChenDMSWYW25,DBLP:conf/eurocrypt/LiuSELBP26} assume that the attacker has access to the exact output logits, which are the direct output of the ReLU network. However, in realistic scenarios, it is unlikely that all logits are exposed to users. For example, in MNIST classification, the network produces a 10-dimensional vector of scores (logits), but the final output is only the predicted class, obtained by taking the argmax over these 10 values. 
Therefore, more recent works have shifted their focus to the \emph{hard-label setting} \cite{DBLP:conf/asiacrypt/ChenDGSWW24,DBLP:conf/eurocrypt/CarliniCHRS25,DBLP:conf/latincrypt/CanalesMartinezS25}, where only the final classification result is visible to the attacker. Chen et al. proposed the first model extraction under this setting, but their method applies only to limited models and requires exponential time complexity \cite{DBLP:conf/asiacrypt/ChenDGSWW24}. 
Later, Carlini et al. proposed a new model extraction that is applicable to multilayer perceptrons (MLPs) and runs in polynomial time \cite{DBLP:conf/eurocrypt/CarliniCHRS25}. 

\begin{figure}[t]
    \centering
    \includegraphics[width=.9\linewidth]{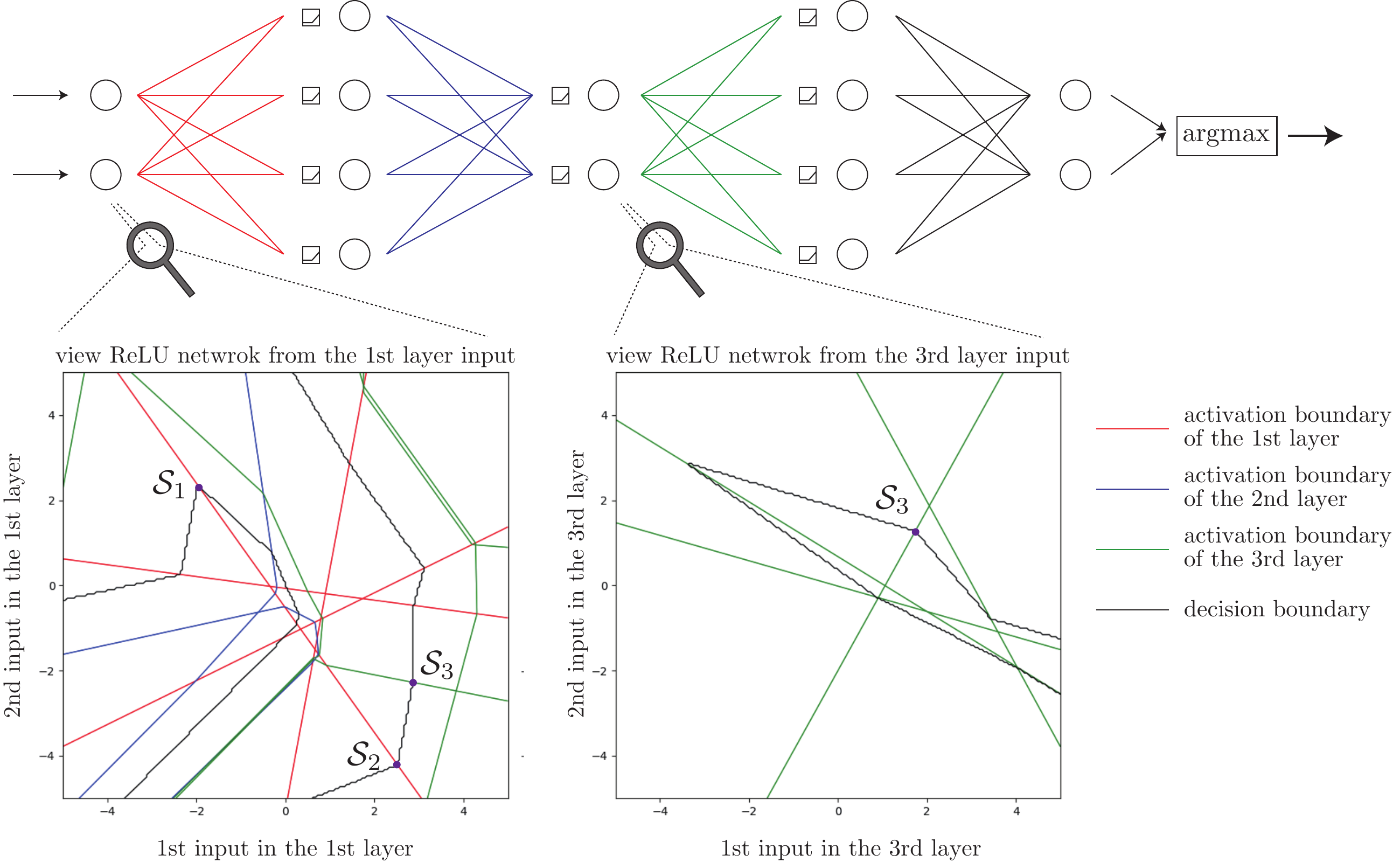}
    \caption{Overview of model extraction with hard-label setting.}
    \label{fig:overview}
\end{figure}
Carlini et al.'s attack inserts sufficiently small differences $\bm \delta$ and detects a point, called an \emph{\intersectionpoint}, where the network's local linear behavior changes. It also detects an affine subspace, called an \emph{\intersectionspace}.\footnote{Carlini et al. used different terms, \emph{dual point} and \emph{dual space}, for \emph{\intersectionpoint} and \emph{\intersectionspace}, respectively. Here, ``dual'' refers to points/space on both activation and decision boundaries, and it differs from the standard notion of a dual space in mathematics. Therefore, to avoid misinterpretation, we use the term \intersectionpoint/\intersectionspace instead.}
\cref{fig:overview} provides a high-level view of why model extraction is feasible. 
Owing to the piecewise linearity of a ReLU network, the input space is partitioned into linear regions. 
Adjacent regions are separated by an \emph{activation boundary}, where exactly one neuron switches between active and inactive. 
Within each region, the activation pattern of all neurons is fixed, and the \emph{decision boundary} (where the output label changes) behaves linearly.
By observing this local linear behavior, an attacker can identify \intersectionpoints and, more generally, \intersectionspaces, where a decision boundary intersects an activation boundary. 
Note that all \intersectionspaces are defined in the input space, e.g., $\mathcal{S}_3$, which lies at the intersection between the decision boundary and the activation boundary of the 3rd layer, is visible from the input. 
Two \intersectionspaces, $\mathcal{S}_1$ and $\mathcal{S}_2$, can lie on the same first-layer activation boundary.
In this case, the attacker can identify the corresponding activation boundary due to its linear nature, thereby recovering the associated secret parameters (weights and biases). 
In contrast, activation boundaries in deeper layers (shown in blue or green) appear nonlinear when viewed from the input space. 
However, as illustrated in the figure, when the network is viewed from an intermediate layer (e.g. the 3rd layer), these boundaries become linear. 
Therefore, once first-layer parameters are recovered, the same strategy can be applied recursively, layer by layer.

%At this \intersectionpoint/\intersectionspace, the attacker can recognize that exactly one neuron out of all neurons has switched between active and inactive states. If this switch occurs in the first layer, the attacker can extract the corresponding weight (a row vector of the weight matrix) and bias (a scalar element of the bias vector). Once all secret weights and biases in the first layer are recovered, the same attack strategy is applied recursively, layer by layer, to extract the entire model. 

\subsection{Our Contribution}
\subsubsection{Is the Hard-Label Extraction Really Polynomial?}
Carlini et al.'s work is remarkably powerful. They claimed that hard-label model extraction is possible in polynomial time in the number of neurons and the network depth. However, their argument implicitly assumes that \intersectionpoints for any target neuron can be collected with a polynomial number of queries. This assumption is reasonable when neuron activation states switch with roughly uniform probability. 

Does this assumption hold in practice? As shown in \cref{fig:overview}, while \intersectionpoints exist between the decision boundary and the activation boundary of the 3rd layer, the ReLU nonlinearity restricts the region that is reachable by the attack. Consequently, if the corresponding \intersectionpoints lie outside the region reachable from the input (e.g., outside the all-positive region), they become inaccessible. Even when reachable \intersectionpoints exist, finding them can be extremely difficult if the reachable intersection region is very narrow. 

To validate our intuition, we conducted experiments using trained networks with high classification accuracy in \cref{subsec:experiment_queries}. We find that inaccessible \intersectionpoints correspond to neurons whose activation states remain fixed, either always active or always inactive, under many queries. Moreover, as depth increases, certain neurons require an exponential number of queries to switch their activation state. We refer to these neurons as \emph{persistent} (always active) and \emph{dead} (always inactive), respectively. 

Persistent/dead neurons have been noted in prior works~\cite{DBLP:conf/eurocrypt/CanalesMartinezCHRSS24,DBLP:conf/eurocrypt/CarliniCHRS25,DBLP:conf/eurocrypt/LiuSELBP26}, but their implications for subsequent layers in model extraction have not been fully analyzed.\footnote{Canales-Martinez et al.~\cite{DBLP:conf/eurocrypt/CanalesMartinezCHRSS24} mention the possibility of persistent neurons but report finding none in their model. Carlini et al.~\cite{DBLP:conf/eurocrypt/CarliniCHRS25} remark that such neurons are rare. Liu et al.~\cite{DBLP:conf/eurocrypt/LiuSELBP26} observe almost such neurons but state that there is no need to identify them individually as absorbable for the recovery of the final layer.} In contrast to prior works, we show that the existence of (almost) persistent neurons constitutes a fundamental challenge to existing hard-label model extraction. When \intersectionpoints cannot be found, the weights and biases of these neurons cannot be recovered. While dead neurons are not critical, as their weights and biases do not influence the final output, the weights of persistent neurons directly affect the behavior of subsequent layers. We show that ignoring these weights inevitably causes non-negligible errors in the extracted model, preventing the attack from proceeding correctly.

Consequently, if almost persistent neurons exist, a successful approach requires an exponential number of queries until they become inactive, which is no longer polynomial time. Moreover, when neurons are truly persistent, model extraction becomes impossible, even with an exponential number of queries.

\subsubsection{Cross-Layer Extraction: Extracting Persistent Neurons.}
To address the challenges posed by persistent neurons, we propose a new technique called \emph{cross-layer extraction}.
When \intersectionpoints cannot be found for (almost) persistent neurons, cross-layer extraction enables their recovery by leveraging interactions across layers. Specifically, it uses \intersectionpoints from neurons in the next layer, which are easier to detect. 
Moreover, this technique can also identify the number of persistent and dead neurons. 
Cross-layer extraction does not necessarily recover the exact target model. 
Instead, it reproduces the original model's outputs as long as neurons classified as persistent or dead remain in those respective states. Since the probability of these states changing is exponentially small, the recovered model produces identical outputs with overwhelming probability. 
Thus, cross-layer extraction reconstructs the best possible model given the information available to the attacker. This stands in contrast to existing approaches, which either fail entirely or incur exponential query cost.

\iflncs
For reproducibility, implementation details and source-code information are provided in the full version at \url{https://eprint.iacr.org/2025/1868}.
\else
For reproducibility, our implementation is available at \url{https://github.com/ECSIS-lab/hard-label-cross-layer-extraction}.
\fi

% If an input is received, where a neuron regarded as persistent during the attack becomes inactive, although it is significantly rare event, the recovered model may produce a different output. Therefore, CrossLayer Extraction reconstructs the best possible model based on the information available to the attacker. This approach stands in contrast to existing model extraction techniques, which either fail entirely or incur exponential cost. 

% 既存は初めはロジットが見える設定を、考えていた。これは、図で言う、出力が全てみえる状態である。現実のDNNほ、数値を、そのまま出力することはない。例えばMNISTの場合、出力次元は、10であり、この10個の値の最大値を持つインデックスが返されてる。ロジットではなく、このargmax の、結果だけを使って、解析する研究がハードラベル設定であり、より現実的だが難しい設定である。AC24では、限られたモデルに対して、指数的計算量をかけて取ることが出来ることが示された。後にEC25では、多項式時間で任意のものが取れると示された。

% 既存研究を総括すると、ReLU型DNNという最も一般的なものにおいては、既に多項式時間で層の深さやニューロンの、数に関わらず、ハードラベルで取れると主張されている。
% 一方で、既存解析は、数多くの仮定に依存している。アルゴリズムが多項式なのは、その仮定が正しい限りである。
% さまざまな仮定のなかには、単純化のためDNNの挙動をランダムと見なしているが、現実は、そこまで単純ではない。

\section{Preliminaries}

% In this section, will organize the notation and explain the model extraction proposed by \citeauthor{carlini2025polynomial}.

\begin{table}[t]
    \centering
    \caption{Notation used in this paper}\label{tab:notation}
    \begin{tabularx}{0.95\textwidth}{cX}
        \hline
        Symbol & Description \\ \hline
        $\bfW^{(\ell)} \in \R^{d_{\ell} \times d_{\ell-1}}$ & weight matrix in the $\ell$th layer \\
        $\bmb^{(\ell)} \in \R^{d_{\ell}}$                   & bias vector in the $\ell$th layer   \\
        $\bmz^{(\ell)} \in \R^{d_{\ell}}$                   & output vector after the $\ell$th layer \\
        % $p_x$ & 入力分布の従う確率密度関数\\
        $f : \R^{d_0} \to \R^{d_L}$                         & ReLU network with $L$ layers      \\
        $f^{(\ell)} : \R^{d_0} \to \R^{d_\ell}$             & ReLU network up to the $\ell$th layer      \\
        $\sigma$                                            & ReLU function \\
        $\bfD^{(\ell)}_{\bm x}$                             & diagonal matrix encoding the $\ell$th-layer activation pattern at $\bm x$ \\
        % $\bfW_{\bm x}, \bm{b}_{\bm x}$                      & $x$における局所アフィン写像の行列とベクトル \\
        % $\hat{y} : \R^{d_L} \to [c]$ & Hard-label モデルのラベリング関数 \\
        % $v_{\mathrm{left}}, v_{\mathrm{right}}$ & 決定境界の法線ベクトル \\
        % \intersectionspace & ReLU境界と決定境界の交差部分 \\
        % Signature & ニューロンの重みの定数倍 \\
        % $\Span \{ v_1, \ldots , v_n \}$ & ベクトル$v_1, \ldots, v_n \in \R^d$が張る空間 \\
        % $V^{\bot}$ & subspace $V \subset \R^d$の直交補空間\\
        $\bms$                                              & \intersectionpoint \\
        $\mathcal{S} (\bms, \bfN)$                          & \intersectionspace\\
        $\bfF_{x}^{(\ell)}$                                 & forward local linear matrix up to the $\ell$th layer for input $\bm x$       \\
        \hline
    \end{tabularx}
\end{table}

In this section, we describe properties of neural networks with ReLU activations. In addition, we present the adversarial capabilities and objectives in the model extraction attacks considered in this paper.
The notation used throughout this paper is summarized in \cref{tab:notation}.

\begin{definition}[ReLU~\cite{DBLP:journals/jmlr/GlorotBB11}]
    A ReLU function, denoted by $\sigma: \R^{d_*} \to \R^{d_*}$, is defined by $\sigma(\bm{x})_i = \max \{ x_i, 0 \}$ for each $1 \le i \le d_*$. 
    \footnote{Although this definition depends on the input dimension, it is customary to denote it by the same symbol $\sigma$, since it applies the same operation component-wise to the input vector.}
\end{definition}

\begin{definition}[ReLU Network]
Let $L \in \mathbb{Z}_{>0}$ be a positive integer.
For $1 \le \ell \le L-1$, we define 
\[
\bm{z}^{(\ell)} = \sigma (\mathbf{W}^{(\ell)} \bm{z}^{(\ell-1)} + \bm{b}^{(\ell)}),
\]
where $\bfW^{(\ell)} \in \R^{d_{\ell} \times d_{\ell-1}}$, $\bmb^{(\ell)} \in \R^{d_{\ell}}$, and $\bmz^{(0)}=\bmx$.
In general, the input and output of the ReLU function are called the pre-activation and the activation, respectively. Here, the pre-activation at layer $\ell$ is given by $\mathbf{W}^{(\ell)} \bm{z}^{(\ell-1)} + \bm{b}^{(\ell)}$, and the activation is denoted by $\bm{z}^{(\ell)}$.
We also define
$\bm{z}^{(L)} = \mathbf{W}^{(L)} \bm{z}^{(L-1)} + \bm{b}^{(L)}$,
as the last layer, where $\bfW^{(L)} \in \R^{d_{L} \times d_{L-1}}$, $\bmb^{(L)} \in \R^{d_{L}}$.
Then, an $L$-layer {\bf ReLU network}\footnote{In this paper, when we refer to a ReLU network, we mean a multi-layer perceptron whose activation function is the ReLU.} is a function $f :  \R^{d_0} \to \R^{d_L}$ such that 
$f(\bmx) = \bmz^{(L)} .$
In particular, we also denote $f^{(\ell)}(\bmx) = \bmz^{(\ell)}$.
A list $\theta = \{(\bfW^{(\ell)}, \bmb^{(\ell)})\}_{\ell=1, \ldots, L}$ is called the parameters of the ReLU network, sometimes also referred to as the weights and biases.
The weight of the $k$th neuron in the $\ell$th layer, denoted by $\bmw_k^{(\ell)} \in \R^{d_{\ell-1}}$, is equal to the transpose of the $k$th row vector of the weight matrix $\bfW^{(\ell)}$.
% We also call the $k$ th row vector of weight matrix $\bfW^{(\ell)}$ the weight of $k$ th neuron in $\ell$ th layer.
Here, we assume that the neural network is a hard-label model.
By using a labeling function $\hat{y} : \R^{d_L} \to [d_L]$, the hard-label model is denoted by $\hat{y} \circ f : \R^{d_0} \to [d_L]$.
\end{definition}

When considering a neural network with hard labels, it can be assumed that the weights of all neurons in each layer are normalized to have unit norm.
This is because, through an appropriate transformation, one can obtain weights that yield the same outputs for any given input.
Furthermore, within each layer, the order of neurons can be rearranged so as to coincide by inserting a suitable permutation matrix in between.

A ReLU network $f$ is locally affine.
Namely, for $\bmx \in \R^{d_0}$, there exist a region $R \subset \R^{d_0}$, a matrix $\bfF_{\bmx}$ and a vector $\bm{\beta}_{\bmx}$ such that $f(\bmy)= \bfF_{\bmx} \bmy + \bm{\beta}_{\bmx}$ for all $\bmy \in R$.
Let us be more specific. 
For each point $\bmx \in \R^{d_0}$, the ReLU network has an activation pattern represented by diagonal matrices $\bfD_{\bmx}^{(1)}, \ldots, \bfD_{\bmx}^{(L-1)}$.
Here, $(\bfD_{\bmx})_{kk}^{(\ell)}=1$ indicates that the $k$th neuron in the $\ell$th layer is activated and $(\bfD_{\bmx})_{kk}^{(\ell)}=0$ means that its value is set to zero by the ReLU.
% For $1 \le \ell \le L$, we set $\bfD^{(\ell)}_{\bmx} \in \R^{d_{\ell}}$ as a diagonal matrix whose diagonal is $\sigma^{\ell}$.
Then, we can describe the forward local linear matrix $\bfF_{\bmx}^{(\ell)}$ and bias $\bm{\beta}_{\bmx}^{(\ell)}$ explicitly as follows:
\begin{align}
\bfF_{\bmx}^{(\ell)} &= \bfD_{\bmx}^{(\ell)}\bfW^{(\ell)} \bfD^{(\ell-1)}_{\bmx} \bfW^{(\ell-1)} \cdots \bfD^{(1)}_{\bmx} \bfW^{(1)}, \\
\bm{\beta}_{\bmx}^{(\ell)} &=  \bfD_{\bmx}^{(\ell)}\bmb^{(\ell)} + \bfD_{\bmx}^{(\ell)}\bfW^{(\ell)} \bfD^{(\ell-1)}_{\bmx} \bmb^{(\ell-1)} + \cdots + \bfD_{\bmx}^{(\ell)}\bfW^{(\ell)} \bfD^{(\ell-1)}_{\bmx}  \cdots \bfD^{(1)}_{\bmx} \bm b^{(1)}.
\end{align}
In particular, we see that $\bfF_{\bmx}^{(L)} = \bfF_{\bmx}$ and $\bm{\beta}_{\bmx}^{(L)} = \bm{\beta}_{\bmx}$, where $\bfD_{\bmx}^{(L)}$ is regarded as the identity matrix $\bfI$.

% \begin{figure}[t]
%     \centering
%     \includegraphics[width=0.8\linewidth]{figures/network-and-boundary.png}
%     \caption{Partitioning of the input space by a ReLU network.}
%     \label{fig:explanationrelunn}
% \end{figure}

% Given an activation pattern, we define the maximal subset of the input space in which this activation pattern remains unchanged as a linear region.
% Linear regions whose activation patterns differ in exactly one entry are adjacent; we refer to the separating surface as an activation boundary (cf. \cref{fig:overview}).
% In the figure, for example, we observe that the input space is partitioned into several linear regions, within each of which the network acts as an affine transformation.
% The red lines represent the activation boundaries where the activation status of the first-layer neurons switches.
% The blue lines indicate the activation boundaries where the second-layer neurons switch their activation.
% Each activation boundary of the second layer bends at the points where it intersects with those of the first layer, but remains linear within each linear region.
% The black line corresponds to the decision boundary.
% The decision boundary bends whenever it crosses an activation boundary.

For a fixed activation pattern, we define the maximal subset of the input space on which the pattern remains unchanged as a linear region.
When two linear regions have activation patterns that differ in exactly one entry, we call them adjacent and define their separating surface as an activation boundary.
As illustrated in \cref{fig:overview}, the input space is partitioned into linear regions, and the network acts as an affine transformation on each region.
The red and blue lines denote activation boundaries of the first and second layers, respectively. The latter bend where they intersect the former, but remain linear within each linear region. The black line denotes the decision boundary, which bends when it crosses an activation boundary.

% An activation boundary is a hyperplane, and when crossing this hyperplane, the activation of a neuron changes. 
% \begin{figure}[t]
%   \centering
%   \begin{subfigure}[t]{0.49\textwidth}
%     \centering
%     \includegraphics[width=0.6\textwidth]{figures/ReLUNN_2-layer.png}
%     \vspace{.5cm}
%   \end{subfigure}
%   \hfill
%   \begin{subfigure}[t]{0.5\textwidth}
%     \centering
%     \includegraphics[width=\textwidth]{figures/ReLUNN_activation_boundaries.png}
%     % \caption{}
%   \end{subfigure}
%   \caption{Partitioning of the input space by a ReLU network}\label{fig:explanationrelunn}
% \end{figure}

% The weight vector corresponding to the neuron serves as the normal vector, which directly reflects the information of the weight matrix of the ReLU network. 
% However, since a hard-label neural network outputs only the class, it is difficult for an attacker to detect this hyperplane. 
% This is where the following \intersectionspace comes into play.

\subsubsection{Adversarial Capabilities and Objectives.} Following \cite{DBLP:conf/eurocrypt/CarliniCHRS25}, we assume the attacker's capabilities and objectives as follows. 
The attacker can observe the outputs when adaptively chosen queries $\bmx$ are provided to an oracle $\mathcal{O}$, where the oracle is the target hard-label DNN $\hat{y} \circ f$. The attacker's objective is to recover parameters $\hat{\theta}$ that are identical to the original model parameters $\theta$ (up to equivalent transformations that do not alter the model's input-output behavior), using sufficiently high-precision floating-point arithmetic. Following prior work, we also make the following assumptions. \textbf{Knowledge of the architecture:} The attacker knows the number of layers and units in the model. \textbf{Full-domain inputs:} The attacker can provide arbitrary inputs in $\mathbb{R}^{d_0}$. \textbf{Precise computations:} The DNN is executed with sufficiently high-precision floating-point arithmetic. \textbf{Fully connected ReLU network:} The target model is a multilayer perceptron (MLP) consisting entirely of fully connected layers, and all activation functions are ReLU.

\section{Model Extraction in EC25}
In this section, we explain the existing model extraction attack proposed by Carlini et al. 
In this attack, the target model is a hard-label ReLU network.
Thus, the only information available to the attacker is the decision boundary at which the classification label changes.
The decision boundary of a ReLU network is composed of piecewise linear regions.
The attacker traverses along this boundary in a straight manner; however, there are locations at which the boundary bends.
We refer to such points as \emph{\intersectionspaces}.
At these points, the activation status of certain neurons in the ReLU network changes, which corresponds to crossing an activation boundary.

 \begin{definition}[Intersection space and \intersectionpoint]
    For a ReLU network, an intersection of an activation boundary and a decision boundary is referred to as an \intersectionspace. Any point contained in an \intersectionspace is called an \intersectionpoint.
     %\footnote{\citeauthor{carlini2025polynomial} refer to these as \emph{dual space} and \emph{dual point}. However, since the dual space of a vector space $V$ is generally understood as the set of linear forms, such terminology can be confusing. Therefore, in this work, we instead refer to them as \intersectionspace and \intersectionpoint, respectively.}
\end{definition}

The attack mainly consists of three steps. 
First, an attacker collects many \intersectionpoints and \intersectionspaces by using the piecewise linearity of the decision boundary. 
Second, a scalar multiple of the weight vector, denoted by \emph{signature}, is recovered using collected \intersectionspaces. 
Finally, the sign of the weight vector and the bias are recovered. 

% The method consists broadly of the following steps: ``collecting \intersectionpoints'', 
% ``recovering the signature with consistency check'', and ``recovering the sign''.

% この攻撃ではhard-labelのReLU networkを対象としている．
% 攻撃者が手掛かりにできるのは分類クラスが変わる決定境界のみである．
% 対象はReLUネットワークなので決定境界はまっすぐな領域になっている．
% 攻撃者はこの境界上をまっすぐ進むが，決定境界が折れ曲がっているとことがある．
% そこを\intersectionspaceと呼ぶ．
% 実はそこでReLUネットワーク内のあるneuronの活性非活性が入れ替わっている，つまりactivation boundaryが存在する．

\begin{figure}[t]
  \centering
  \begin{subfigure}[t]{0.50\textwidth}
    \centering
    \includegraphics[width=\textwidth]{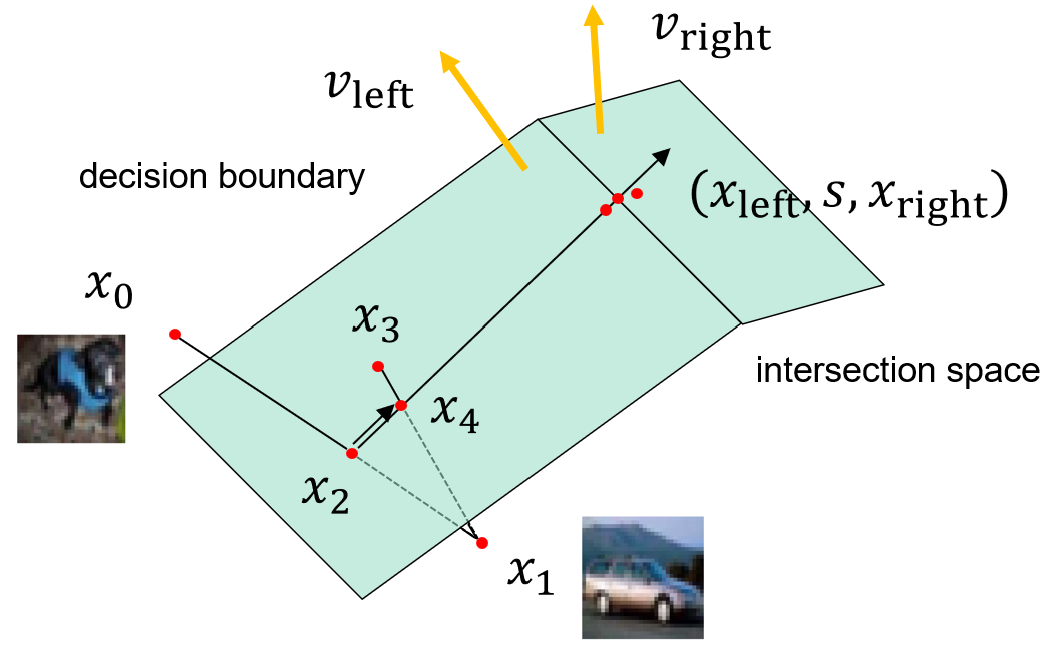}
    \caption{Method for finding intersection spaces (\cref{subsec:collectIS}).}\label{fig:onthedecisionboundary}
  \end{subfigure}
  \hfill
  \begin{subfigure}[t]{0.48\textwidth}
    \centering
    \includegraphics[width=\textwidth]{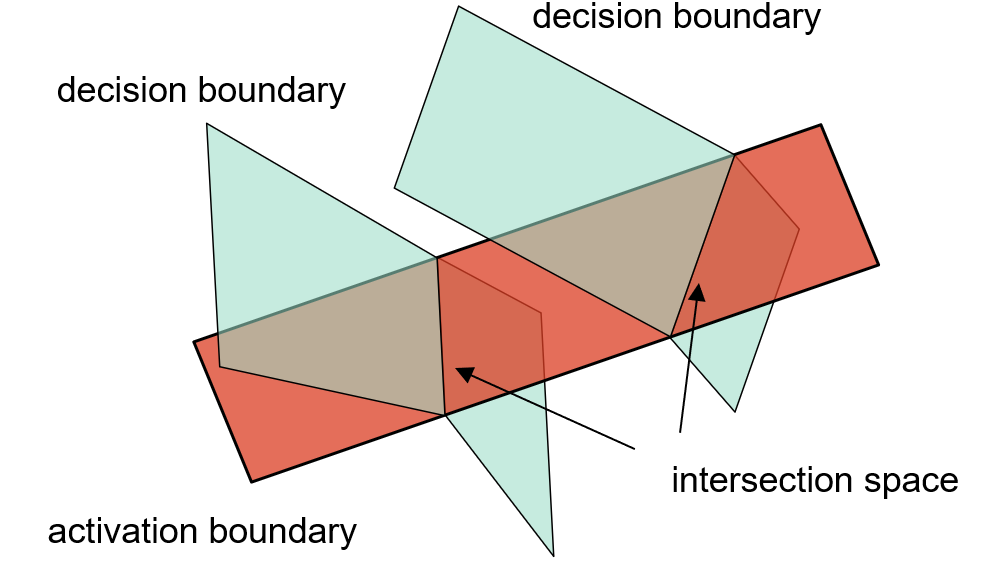}
    \caption{Consistent intersection spaces are contained in the same activation boundary (\cref{subsec:signature}).}\label{fig:sameactivationboundary}
  \end{subfigure}
  \caption{Intersection point search and signature recovery.}
\end{figure}

% 攻撃の全体像は次の通りである．
% まず，復元に不可欠である\intersectionspacesを集められるだけ集めていく．
% 集められたら，浅い層から順番に復元していく．
% まず1層目の各ニューロンのSignatureを復元していく（Definition~\ref{def:signature}）．
% これは向きを除いたニューロンの重みになっている．
% 次にSign recoveryで符号を復元する．
% これにより1層目が完全に復元された．
% 1層目の情報を使いながら2層目のニューロンのSignatureをとり，signを復元し順番に重みを復元していく．

\subsection{Collecting Intersection Points}\label{subsec:collectIS}
We first collect the \intersectionpoints, as outlined in \cref{fig:onthedecisionboundary}.
% First, we collect the \intersectionpoints.
% An outline of this procedure is shown in \cref{fig:onthedecisionboundary}.

\paragraph{Step 1: Find a point $\bmx_2$ on the decision boundary.}
We sample two random inputs $\bmx_0, \bmx_1 \in \R^{d_0}$ with $\hat{y}(f(\bmx_0)) \neq \hat{y}(f(\bmx_1))$ and use binary search between them to find a point $\bmx_2$ on the decision boundary.
%We call this point $\bmx_{\mathrm{left}}$.
Here, $\bmx_0$ and $\bmx_1$ are drawn from an appropriate distribution (e.g., a Gaussian distribution).

\paragraph{Step 2: Find another point $\bmx_4$ on the decision boundary.}
We move away from $\bmx_2$ in a random direction to obtain a new point $\bmx_3$. % sufficiently close to the same decision boundary. 
Without loss of generality, assume that $\hat{y}(f(\bmx_3)) = \hat{y}(f(\bmx_0))$; otherwise swap $\bmx_0$ and $\bmx_1$.
A binary search between $\bmx_1$ and $\bmx_3$ then yields another point $\bmx_4$ on the decision boundary.
%  because the predicted label of $\bmx_3$ should be equivalent to that of either $\bmx_1$ or $\bmx_0$.
% We can find a point $\bmx_4$ on the decision boundary by a binary search between $\bmx_1$ and $\bmx_3$.

\paragraph{Step 3: Move along the decision boundary until an \intersectionpoint $\bms \in \mathbb{R}^{d_0}$ is found.}
Set $\Delta \bmx := \bmx_4-\bmx_2$ and move along the boundary via $\bmx_{\alpha}:=\bmx_2+\alpha\Delta\bmx$.
For sufficiently small $\alpha$, $\bmx_{\alpha}$ remains on the decision boundary.
Once $\bmx_{\alpha}$ leaves the boundary, the path has crossed an activation boundary.
We denote the crossing point by $\bms$ and call it an \intersectionpoint.

% After some progression, $\bmx_{\alpha}$ is no longer located on the decision boundary.
% At this point, the path has crossed an activation boundary of the ReLU network.
% The point at which this crossing occurs is referred to as an intersection point and is denoted by $\bms$.

\paragraph{Step 4: Re-locate the decision boundary.}
After crossing the activation boundary, we project the point back onto the decision boundary to obtain $\bmx_{\mathrm{right}}$.
Let $\bmx_{\mathrm{left}}$ denote the point $\bmx_{\alpha}$ immediately before the crossing.
Then, we have collected a triplet of points: $(\bmx_{\mathrm{left}}, \bms, \bmx_{\mathrm{right}})$.
% After crossing the activation boundary, the point is projected back onto the decision boundary.
% This projected point is referred to as $\bmx_{\mathrm{right}}$, while the point $\bmx_{\alpha}$ immediately before crossing the activation boundary is referred to as $\bmx_{\mathrm{left}}$.
% Then, we have collected a triplet of points: $(\bmx_{\mathrm{left}}, \bms, \bmx_{\mathrm{right}})$.

\paragraph{Step 5: Find the normal vectors $\bmv_{\mathrm{left}}$ and $\bmv_{\mathrm{right}}$.}
Let $\bmx \in \R^{d_0}$ be a point on the decision boundary, and let $\bmx'=\bmx+\bm{\epsilon}$ be a slight perturbation off the boundary.
For each standard basis vector $\bme_i$, find $\alpha_i$ such that $\bmx'+\alpha_i\bme_i$ lies on the decision boundary.
Then normalizing $(1/\alpha_1,\ldots,1/\alpha_{d_0})^\top$ yields a unit normal vector of the decision boundary at $\bmx$.
Applying the same procedure to $\bmx_{\mathrm{left}}$ and $\bmx_{\mathrm{right}}$ yields $\bmv_{\mathrm{left}}$ and $\bmv_{\mathrm{right}}$.

% Let $\bmx \in \R^{d_0}$ be a point on the decision boundary.
% Let $\bmx' = \bmx + \bm{\epsilon}$, so that the point is slightly displaced from the boundary.
% Using the standard basis vectors $\bme_1, \ldots, \bme_{d_0}$, we find scalars $\alpha_i$ such that $\bmx' + \alpha_i \bme_i$ lies on the decision boundary for each direction.
% At this point, the normal vector $\bmv \in \R^{d_0}$ of the decision boundary is aligned with $(1/\alpha_1, \ldots, 1/\alpha_{d_0})^{\top}$.
% By normalization, we obtain a unit-length normal vector.
% For each point $\bmx_{\mathrm{left}}$ and $\bmx_{\mathrm{right}}$, we can thus derive the corresponding unit normal directions $\bmv_{\mathrm{left}}$ and $\bmv_{\mathrm{right}}$.

\paragraph{Step 6: Describe \intersectionspace.}

We can express an \intersectionspace $\mathcal{S}$ of the \intersectionpoint $\bms$ as $\bms + \Span \{ \bmv_{\mathrm{left}}, \bmv_{\mathrm{right}} \}^{\perp} \subset \R^{d_0}$.
Set an orthonormal basis of $\Span \{ \bmv_{\mathrm{left}}, \bmv_{\mathrm{right}} \}^{\perp}$ as $\bmn_1, \ldots, \bmn_{d_0-2} \subset \R^{d_0}$.
Let $\bfN \in \mathbb{R}^{d_0 \times (d_0-2)}$ denote the matrix obtained by concatenating $\bmn_1, \ldots, \bmn_{d_0-2}$.
Then, the intersection space $\mathcal{S}$ can be represented by $(\bms, \bfN) \in \mathbb{R}^{d_0} \times \mathbb{R}^{d_0 \times (d_0-2)}$, where $\bms$ is an arbitrary point in $\mathcal{S}$ and $\bfN$ is the matrix whose columns form an orthonormal basis for the linear subspace that specifies its direction.

\paragraph{Step 7: Repeat these steps.}
By sufficiently repeating the above six steps, we collect a large number of intersection spaces $\mathcal{S}_1, \ldots, \mathcal{S}_N$.

\subsection{Recovering the Signature}\label{subsec:signature}

% neuronのweightは対応するactivation boundaryの法線ベクトルになっている．
% 複数のintersectionspaceが同じactivation boundaryに含まれるときas shown in \cref{fig:sameactivationboundary}，そのactivation boundaryを復元することができる．
% その法線ベクトルを計算することで，対応するニューロンの定数倍を得ることができる．
% これをsignatureと呼ぶ．
% このsubsectionではsignatureの復元方法を説明する．
The weight of a neuron corresponds to the normal vector of its associated activation boundary.
When multiple intersection spaces are contained in the same activation boundary, as shown in \cref{fig:sameactivationboundary}, the activation boundary can be reconstructed.
By computing its normal vector, we obtain a scalar multiple of the corresponding neuron's weight, which we refer to as its \emph{signature}.
In this subsection, we describe the procedure for recovering the signature.

% Using the large collection of intersection spaces, we reconstruct the weight vectors corresponding to the weights of neurons, up to their signs.
% While an \intersectionspace itself has dimension $d_0 - 2$, , the presence of two \intersectionspaces lying on the same activation boundary allows for the reconstruction of its defining equation.

\begin{definition}[Signature]\label{def:signature}
    Let $\bmw_k^{(\ell)} \in \R^{d_{\ell-1}}$ be the weight of the $k$th neuron in the $\ell$th layer, which is equal to the transpose of $k$th row vector of the weight matrix $\bfW^{(\ell)}$.
    The signature of the neuron is a vector $\alpha \cdot \bmw_k^{(\ell)} \in \R^{d_{\ell-1}}$ for some $\alpha \in \R\backslash \{ 0 \}$. 
    % $\alpha \cdot (\bfW^{(l)}_{k1}, \bfW^{(l)}_{k2}, \ldots, \bfW^{(l)}_{kd_{l-1}})^{\top}$.
\end{definition}

% \paragraph{Observation.}

% Signatureの復元は大きく2段階で行われる．
% 1つめがconsistency check $\texttt{IsConsistent}()$，2段目が$\texttt{RecoverSignature}()$である．
% consistency checkでは2つの\intersectionspacesが同一のactivation boundaryに含まれているものなのかを確認する．
% 同一のactivation boundaryに含まれている\intersectionspacesが十分に集まったら，それらの情報からactivation boundaryを復元し，Signature recoverを行う．

The recovery of the signature is carried out in two main stages.
The first stage is the consistency check, $\texttt{IsConsistent}()$ (\cref{alg:pre-Consistent}), and the second is $\texttt{RecoverSignature}()$ (\cref{alg:pre-signaturerec}).
In the consistency check, we verify whether two intersection spaces belong to the same activation boundary.
Once a sufficient number of intersection spaces belonging to the same activation boundary have been collected, the activation boundary is reconstructed from this information, and the signature recovery is performed.
Note that we assume that $\bfF_{\bmx}^{(0)} = \bfI_{d_0}$ for all $\bmx \in \R^{d_0}$, where $\bfI_{d_0}$ is the identity matrix of size $d_0$.
\begin{algorithm}[tb]
    \caption{\texttt{IsConsistent}($\mathcal{S}_1$, $\mathcal{S}_2$, $\{ \bfW^{(1)}, \ldots, \bfW^{(\ell-1)}\}$)} \label{alg:pre-Consistent}
    \algrenewcommand\algorithmicrequire{\textbf{Input:}}
    \algrenewcommand\algorithmicensure{\textbf{Output:}}
    \begin{algorithmic}[1] 
        \Require $\mathcal{S}_1, \mathcal{S}_2$: \intersectionspaces, $\bfW^{(1)}, \ldots, \bfW^{(\ell-1)}$:  recovered weights
        \Ensure False if the $\mathcal{S}_1, \mathcal{S}_2$ are not on the same activation boundary, otherwise True.
        \State Let $\bfN_1, \bfN_2$ be basis matrices and $\bms_1, \bms_2$ \intersectionpoints, corresponding to $\mathcal{S}_1, \mathcal{S}_2$, respectively
        % \If{$\mathbb{W} = \emptyset$}
        % \Comment{The target is the first layer.}
        % \If{$\dim \Span \{ \bfN_1, \bfN_2 \} < d_0$}
        % \State \textbf{return} True
        % \Else
        % \State\textbf{return} False
        % \EndIf
        % \Else
        \State Let $\bfF^{(\ell-1)}_{\bms_1}, \bfF^{(\ell-1)}_{\bms_2}$ denote the forward local linear matrices.
        \State Let $\nu$ be the number of neurons active in either $\bms_1$ or $\bms_2$.
        \Comment{If $\ell=1$, then $\nu = d_0$.}
        \If{$\dim( \mathrm{Im}(\bfF^{(\ell-1)}_{\bms_1} \bfN_1) + \mathrm{Im}(\bfF^{(\ell-1)}_{\bms_2} \bfN_2) ) < \nu$}
        \State \textbf{return} True
        \Else
        \State \textbf{return} False
        \EndIf
        % \EndIf
    \end{algorithmic}
\end{algorithm}

\begin{algorithm}[tb]
    \caption{\texttt{RecoverSignature}($\{ \mathcal{S}_1, \ldots, \mathcal{S}_m\}$, $\{ \bfW^{(1)}, \ldots, \bfW^{(\ell-1)}\}$)} \label{alg:pre-signaturerec}
    \algrenewcommand\algorithmicrequire{\textbf{Input:}}
    \algrenewcommand\algorithmicensure{\textbf{Output:}}
    \begin{algorithmic}[1] 
        \Require $\mathcal{S}_1, \ldots, \mathcal{S}_m$: consistent \intersectionspaces corresponding to the $k$th neuron in the $\ell$th layer, $\bfW^{(1)}, \ldots, \bfW^{(\ell-1)}$:  recovered weights
        \Ensure $\alpha \cdot \bmw_{k}^{(\ell)} \in \R^{d_{\ell-1}}$: signature of the neuron
        \State Let $\bfN_1, \ldots, \bfN_m$ be basis matrices and $\bms_1, \ldots, \bms_m$ \intersectionpoints, corresponding to $\mathcal{S}_1, \ldots, \mathcal{S}_m$, respectively
        % \If{$\mathbb{W} = \emptyset$}
        % \Comment{The target is the first layer.}
        % \State Find a vector $w \in \R^{d_0}$ such that $\Span \{ w, \bfN_1, \ldots , \bfN_m \} = \R^{d_0}$
        % \State \textbf{return} $w$
        % \Else
        % \Comment{The target is a deeper layer.}
        \State Set $\bfF^{(\ell-1)}_{\bms_1}, \ldots, \bfF^{(\ell-1)}_{\bms_m}$ be forward local linear matrices.
        \If{$\dim (\mathrm{Im} (\bfF^{(\ell-1)}_{\bms_1} \bfN_1) + \cdots + \mathrm{Im} (\bfF^{(\ell-1)}_{\bms_m} \bfN_m))$ < $d_{\ell-1}-1$}
        \State \textbf{return} $\bot$
        \Comment{Insufficient number of \intersectionspaces.}
        \Else
        \State Find $w \in \R^{d_{\ell-1}}$ such that $\Span \{ w, \bfF^{(\ell-1)}_{\bms_1} \bfN_1, \ldots, \bfF^{(\ell-1)}_{\bms_m} \bfN_m \} = \R^{d_{\ell-1}}$
        \State \textbf{return} $w$
        \EndIf
        % \EndIf
    \end{algorithmic}
\end{algorithm}

\paragraph{Case 1: The first layer.}
We begin by considering the case of the first layer.
Let $\mathcal{S}_1$ and $\mathcal{S}_2$ be \intersectionspaces, and $\bfN_1, \bfN_2$ corresponding bases.
If the two \intersectionspaces are contained in the same activation boundary, it holds that
$\dim \mathrm{span} \{ \bfN_1, \bfN_2 \} < d_0$.
% ここで，$\mathrm{span} \{ \bfN_1, \bfN_2 \}$とは$\bfN_1, \bfN_2$に入っている基底を合わせたもので貼られる$\R^{d_0}$内の部分線形空間である．
Here, $\mathrm{span}\{\bfN_1, \bfN_2\}$ denotes the subspace of $\mathbb{R}^{d_0}$ spanned by the combined bases contained in $\bfN_1$ and $\bfN_2$.
The converse of the above statement also holds.

Suppose that a set of consistent intersection spaces has been collected.
For two intersection spaces $\mathcal{S}_1$ and $\mathcal{S}_2$ that are confirmed to be consistent, we then perform $\texttt{RecoverSignature}()$.
Here, the unique missing dimension corresponding to the defining equation of the target activation boundary represents the signature of the target neuron.

By repeating this procedure sufficiently many times, we obtain the complete collection of signatures of the first layer.

\paragraph{Case 2: Deeper layers.}
Suppose that the weights $\bfW^{(1)}, \ldots, \bfW^{(\ell-1)}$ up to the $(\ell-1)$st layer have been recovered.

We first perform the consistency check.
Let $\mathcal{S}_1$ and $\mathcal{S}_2$ be \intersectionspaces.
Denote the corresponding \intersectionpoints and basis matrices by $\bms_1, \bms_2$ and $\bfN_1, \bfN_2$, respectively.
Let $\bfF^{(\ell-1)}_{\bms_1}, \bfF^{(\ell-1)}_{\bms_2}$ be the forward local linear matrices.
% Define  as $\bfF^{(\ell-1)}_{\bms_1}$.
% In particular, for the point $\bms_1$, letting the activation patterns up to the $(\ell-1)$st layer be $\bfD^{(1)}_{\bms_1}, \ldots, \bfD^{(\ell-1)}_{\bms_1}$, this implies that
% \[
% \bfF_{\bms_1}^{(\ell-1)} = \bfD^{(\ell-1)}_{\bms_1} \bfW^{(\ell-1)} \cdots \bfD^{(1)}_{\bms_1} \bfW^{(1)}.
% \]
% $\bfF_{\bms_2}^{(\ell-1)}$ is defined in the same manner.
If $\mathcal{S}_1$ and $\mathcal{S}_2$ correspond to the same neuron, the following relation holds:
\[
\dim( \mathrm{Im}(\bfF^{(\ell-1)}_{\bms_1} \bfN_1) + \mathrm{Im}(\bfF^{(\ell-1)}_{\bms_2} \bfN_2) ) < \nu.
\]
where $\nu$ is the number of neurons active in either $\bms_1$ or $\bms_2$.
The converse also holds.
We next consider the recovery of the signature.
Suppose that, for each neuron, a sufficiently large number of consistent intersection spaces $\mathcal{S}_1, \ldots, \mathcal{S}_m$ have been collected.
Then, the signature is recovered by $\texttt{RecoverSignature}()$ (\cref{alg:pre-signaturerec}).

\subsection{Recovering the Sign}
Signature recovery results in a scalar multiple of the weight vector. Since there is an equivalent neural network that uses normalized weight vectors, the scalar scale is not important. However, the sign (i.e., the direction) of the weight vector is still important. 
The sign affects the input to the ReLU function and, ultimately whether the neuron is active or inactive. 

Let $\bm s$ be an \intersectionpoint associated with the $k$th neuron in the $\ell$th layer. 
At this point, the pre-activation of the corresponding neuron is 0. 
Let $\bm w_k^{(\ell)}$ be its weight vector, and suppose we have already obtained a signature $\alpha \bm w_k^{(\ell)}$ and the forward local linear matrix $\mathbf{F}^{(\ell-1)}_{\bm s}$. 
We then take a sufficiently small displacement $\bm \delta = \mathbf{F}_{\bm s}^{(\ell-1)\top} (\alpha \bm w_k^{(\ell)})$, which represents the normal vector in the input space to the activation boundary of the $k$th neuron in the $\ell$th layer.
Then, 
\begin{align}
    \bm w_k^{(\ell)\top} \left( f^{(\ell-1)}(\bm s + \bm \delta) - f^{(\ell-1)}(\bm s) \right) &= \bm w_k^{(\ell)\top} \mathbf{F}_{\bm s}^{(\ell-1)} \mathbf{F}_{\bm s}^{(\ell-1)\top}(\alpha \bm w_k^{(\ell)}) \\
    &= \alpha \left\| \mathbf{F}_{\bm s}^{(\ell-1)\top} \bm w_k^{(\ell)} \right\|^2.
\end{align}
% \begin{align}
%     \bm w_k^{(\ell)\top} \bigl( f^{(\ell-1)}(\bm s + \bm \delta) \DIFadd{- f^{(\ell-1)}(}\bm \DIFadd{s)}\bigr)
%     & = \bm w_k^{(\ell)\top} \mathbf{F}_{\bm s}^{(\ell-1)} {\delta }\\
%     & = \alpha  \left\| \mathbf{F}_{\bm s}^{(\ell-1)\top} \bm w_k^{(\ell)}  \right\|^2.
% \end{align}
When $\alpha > 0$, adding and subtracting $\bm \delta$ make the pre-activation positive and negative, respectively. 
On the other hand, when $\alpha < 0$, adding and subtracting $\bm \delta$ make the pre-activation negative and positive, respectively. 
Sign recovery is based on the expectation that moving to the positive side would switch the state of neurons in deeper layers more frequently because the ReLU suppresses negative pre-activation. 

In this paper, we mainly focus on the difficulty of finding \intersectionpoints. Therefore, we omit further details about the sign recovery, and please refer to the original paper \cite{DBLP:conf/eurocrypt/CarliniCHRS25}.

% \subsection{Soft-Label Cryptanalytic Model Extraction.}
% % 

%%%%%%%%%%%%%%%%%%%%%%%%%%%%%%%%%%%%%%%%%%%%%%%%%%%
%%%%%%%%%%%%%%%%%%%%%%%%%%%%%%%%%%%%%%%%%%%%%%%%%%%

\section{Model Extraction in EC25 Is Exponentially Hard} \label{sec:persistent}

Carlini et al. proposed an algorithm that estimates all parameters of a MLP with ReLU activations using a number of queries polynomial in the number of neurons. In this section, we show that due to the presence of neurons that are always active (which we refer to as \textit{persistent neurons}), it can be difficult to recover all neurons with only polynomially many queries. We first provide a definition of persistent neurons in \cref{subsec:persistent_neuron}, and then, in \cref{subsec:experiment_queries}, empirically demonstrate using models trained on MNIST and Fashion-MNIST (FMNIST) that when Gaussian-distributed vectors are used as inputs, there exist neurons whose activation states are practically impossible to switch without employing an exponentially large number of queries with respect to the depth of the network. In \cref{subsec:error_induced_by_persistent_neuron}, we further show that the existence of persistent neurons introduces errors in the recovery of deeper-layer neurons, even if computations are assumed to be carried out with arbitrary precision. In \cref{sec:bypass}, we present an algorithm that successfully recovers neurons even in the presence of persistent neurons.

\subsection{Dead Neurons and Persistent Neurons} \label{subsec:persistent_neuron}

The algorithm proposed by Carlini et al. requires at least two \intersectionpoints for each neuron to recover the corresponding weights. In their method, input samples $\{\bmx_i\}_{i=1}^N$ are drawn from a suitable input distribution $p_{\bmx}$ (e.g., a Gaussian distribution) and fed into the model. For each pair of inputs $\bmx_i$ and $\bmx_j$ that produce different classification outcomes, a binary search is conducted to estimate the decision boundary, which is then used to locate an \intersectionpoint. Consequently, if the distribution $p_{\bmx}$ does not yield samples that fall into both the active and inactive regions of each neuron, the attack becomes infeasible. This observation suggests that in Carlini et al.'s algorithm, neurons with extremely low activation or inactivation probabilities may hinder the attack.

To capture such cases, we define neurons with low activation or inactivation probabilities as dead neurons and persistent neurons, respectively:
\begin{definition}[Dead neuron and persistent neuron]
Let $p_{\bmx}$ be a probability density function over the input space $\mathbb{R}^{d_0}$, and let $\varepsilon \in [0, 1]$ be a positive real number. A neuron whose activation probability is at most $\varepsilon$ (i.e., for layer $\ell$ and neuron index $i$,
$\Pr(\sigma(\bmw^{(\ell)\top}_i  f^{(\ell-1)}(\bmx) + b_i\vp{\ell}) > 0) \leq \varepsilon)$
is called a $(p_{\bmx}, \varepsilon)$-dead neuron. Similarly, a neuron whose inactivation probability is at most $\varepsilon$ (i.e.,
$\Pr(\sigma(\bmw^{(\ell)\top}_i f^{(\ell-1)}(\bmx) + b_i\vp{\ell}) = 0) \leq \varepsilon)$
is called a $(p_{\bmx}, \varepsilon)$-persistent neuron.
\end{definition}
When the input distribution $p_{\bmx}$ is clear from context, we simply refer to these as $\varepsilon$-dead neurons and $\varepsilon$-persistent neurons. Furthermore, when $\varepsilon$ itself is also fixed, we omit it and use the terms dead neuron and persistent neuron.

If the adversary's query budget is limited to $N$, then neurons with $\varepsilon < 1/N$ are unlikely to switch between active and inactive states within the available queries. In such cases, Carlini et al.'s algorithm will, with high probability, fail to recover the weights of these neurons. This stands in contrast to the original assumption of their algorithm, where it is expected that the query budget is sufficient to change the activation state of every neuron. Failure to estimate the weights of dead or persistent neurons can propagate errors to the estimation of neurons in subsequent layers. In particular, if the attack proceeds under the assumption that the weights of such neurons are zero, dead neurons pose no issue since their outputs vanish. However, persistent neurons can have a significant impact on the final output, leading to severe consequences. While the effects will be discussed in \cref{subsec:error_induced_by_persistent_neuron}, we first empirically investigate the existence of dead and persistent neurons in each layer of neural networks in the next subsection.

\subsection{Exponential Queries Required for Persistent Neurons} \label{subsec:experiment_queries}
In this subsection, we empirically demonstrate that, in trained models, there exist neurons whose activation or inactivation probabilities decrease exponentially with respect to the network depth. Moreover, we show that there is a strong proportional relationship between the probability of neuron state switching and the number of \intersectionpoints discovered by the algorithm of Carlini et al. These experimental results indicate that, as the depth increases, recovering the weights of dead or persistent neurons becomes exponentially difficult.

\paragraph{Number of dead and persistent neurons when varying $\varepsilon$.}
We begin by describing the experimental setup. The datasets used were MNIST and Fashion-MNIST (FMNIST), and we trained a 20-layer MLP. Each hidden layer consisted of 256 units, with weights initialized using Kaiming normal initialization~\cite{DBLP:conf/iccv/HeZRS15} and all biases set to zero. The optimizer was Adam~\cite{DBLP:journals/corr/KingmaB14} with a learning rate of $1 \times 10^{-5}$, and the number of training epochs was set to 30.

\begin{figure}[t]
    \begin{subfigure}[t]{0.32\textwidth}
        \centering
        \includegraphics[width=\linewidth]{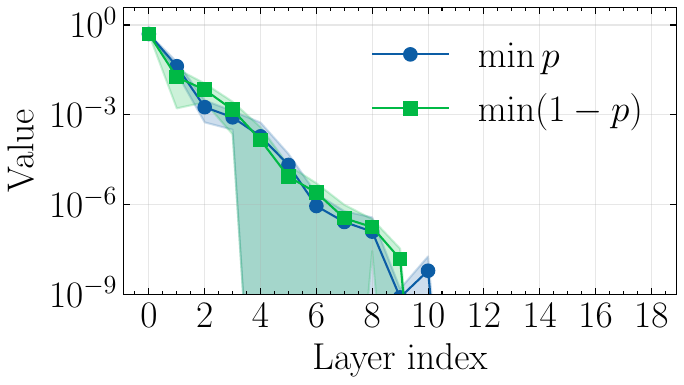}
        \caption{Untrained model.}
        \label{subfig:no_training}
    \end{subfigure}
    \begin{subfigure}[t]{0.32\textwidth}
        \centering
        \includegraphics[width=\linewidth]{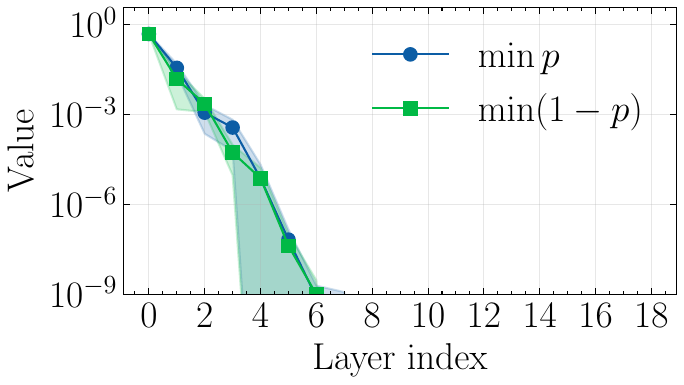}
        \caption{MNIST.}
        \label{subfig:prob_MNIST}
    \end{subfigure}
    \begin{subfigure}[t]{0.32\textwidth}
        \centering
        \includegraphics[width=\linewidth]{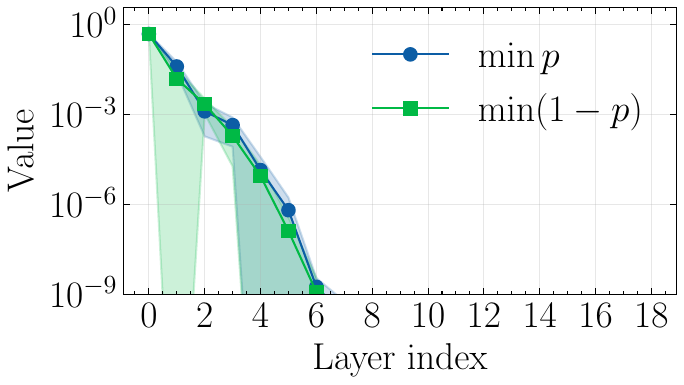}
        \caption{FMNIST.}
        \label{subfig:prob_FMNIST}
    \end{subfigure}
    \caption{Layer-wise minimum neuron activation (or inactivation) probabilities.}
    \label{fig:prob}
\end{figure}

\begin{figure}[t]
    \begin{subfigure}[t]{0.32\textwidth}
        \centering
        \includegraphics[width=\linewidth]{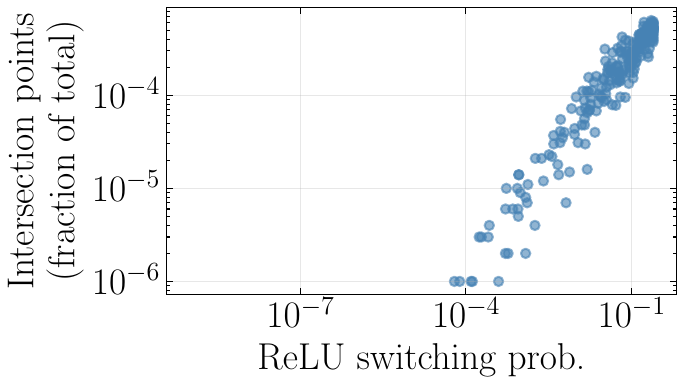}
        \caption{Untrained model.}
        \label{subfig:no_training_switching}
    \end{subfigure}
    \begin{subfigure}[t]{0.32\textwidth}
        \centering
        \includegraphics[width=\linewidth]{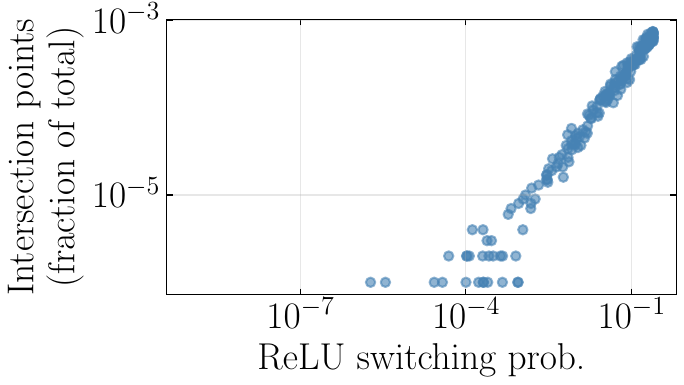}
        \caption{MNIST.}
        \label{subfig:prob_MNIST_switching}
    \end{subfigure}
    \begin{subfigure}[t]{0.32\textwidth}
        \centering
        \includegraphics[width=\linewidth]{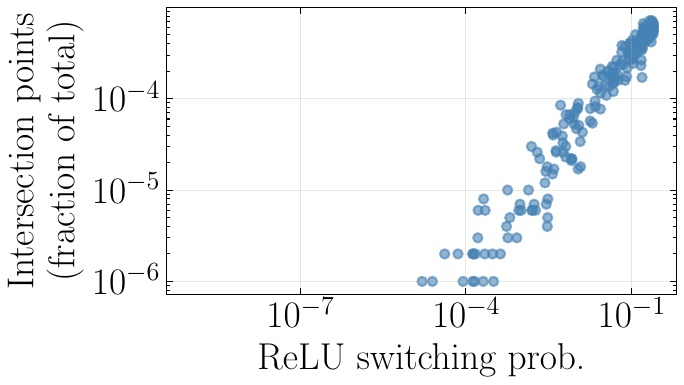}
        \caption{FMNIST.}
        \label{subfig:prob_FMNIST_switching}
    \end{subfigure}
    \caption{Switching probability versus the proportion of discovered \intersectionpoints in the 10th layers of the trained MLPs.}
    \label{fig:dual_vs_toggle_prob}
\end{figure}
%\begin{figure}[t]
%    \centering
%    \includegraphics[width=\linewidth]{figures/all_layers_grid_4x5.pdf}
%    \caption{Switching probability versus the proportion of discovered \intersectionpoints in each layer of the MLP trained on MNIST.}
%    \label{fig:dual_vs_toggle_prob}
%\end{figure}
Let $p_i^{(\ell)}$ denote the activation probability of the $i$th neuron in layer $\ell$. For each layer, we computed both the minimum activation probability (i.e., $\min_i p_i^{(\ell)}$) and the minimum inactivation probability (i.e., $\min_i (1-p_i^{(\ell)})$). These results are shown in \cref{fig:prob}. For each dataset, we trained five models with different random seeds, and we report the mean and standard deviation of the minimum probability values across models.  The average test accuracies of the trained models are 94.7\% for MNIST and 84.8\% for FMNIST. For reference, we also show results for an untrained, randomly initialized model in \cref{subfig:no_training}, where activation patterns were measured over $10^9$ input points sampled from a Gaussian distribution with mean 0 and variance 1. As the figures indicate, there exist neurons whose activation or inactivation probabilities become exponentially small with increasing network depth. When an empirical probability is zero under the finite number of samples, the corresponding point is not displayed on the logarithmic scale; such a missing point means that the probability is below our experimental resolution, not that the true probability is mathematically zero.

Furthermore, we explored $10^6$ \intersectionpoints and examined the relationship between the discovery rate of \intersectionpoints for each neuron, which is defined as the number of \intersectionpoints discovered for that neuron divided by the total number of explored \intersectionpoints, and the corresponding switching probability $2 p_i^{(\ell)} (1-p_i^{(\ell)})$. \cref{fig:dual_vs_toggle_prob} plots these quantities for the 10th layer of the trained models, where the x-axis shows the switching probability and the y-axis shows the discovery rate. 
\iflncs
% The experimental results for all layers are provided in the full version \cite{DBLP:journals/iacr/ItoMT25}. 
The experimental results for all layers are provided in the full version at \url{https://eprint.iacr.org/2025/1868}. 
\else
The experimental results for all the layers are shown in \cref{fig:all_layer_MNIST_0,fig:all_layer_MNIST_30,fig:all_layer_FMNIST_30}. 
\fi
The results clearly demonstrate a strong proportional relationship between them. In other words, neurons that are sufficiently close to being dead or persistent (i.e., with extremely small switching probability $\varepsilon$) are unlikely to yield \intersectionpoints. Taken together, these results indicate that, in deeper layers, discovering \intersectionpoints for neurons with extremely high activation or inactivation probabilities becomes exponentially difficult.

\subsection{Analyzing Estimation Errors Induced by Persistent Neuron} \label{subsec:error_induced_by_persistent_neuron}

In the previous subsection, we empirically observed that recovering persistent neurons becomes exponentially harder as the network depth increases. In other words, as layers grow, an adversary may need exponentially more queries to recover all neurons. Consequently, to run the algorithm of Carlini et al. in polynomial time with respect to the number of neurons, the estimation of dead or persistent neurons with sufficiently small $\varepsilon$ must be disregarded.

In the following subsections, we analyze the error that arises when estimating neurons in deeper layers under the condition that at least one persistent neuron has not been recovered in an earlier layer. We show that this leads to an additional unavoidable error in neuron signature estimation. This implies that if even a single persistent neuron is missed in an earlier layer, the resulting estimation error propagates to subsequent layers and may cause the attack to fail. Here, we assume that consistency checks can still be performed.\footnote{In practice, the presence of persistent neurons may also break consistency. However, in this analysis we assume that consistency checks remain valid and focus only on the additional noise introduced into neuron recovery. If neuron recovery fails, it ultimately indicates that existing attacks are unsuccessful.}

%To this end, we first recall the original neuron recovery algorithm presented by Carlini et al. and explain how the presence of persistent neurons in earlier layers affects recovery. Next, we analyze the error introduced by persistent neurons under simplified conditions, showing that an error appears. Finally, in \cref{subsec:error_experiment}, we experimentally confirm the impact of persistent neurons.  

\subsubsection{Effect of Persistent Neurons in Existing Methods.}
We begin by recalling the neuron estimation algorithm and examining the influence of the presence of a persistent neuron on the estimation. In particular, when a persistent neuron exists, the attacker is forced to treat it as dead, and therefore cannot recover the corresponding coordinate. Consequently, the signature recovery problem in the next layer reduces to an estimation problem with that coordinate removed. We formalize this relationship below. 

Suppose an adversary has extracted $m$ \intersectionpoints $(\bms_j)_{j\in [m]}$ corresponding to neuron $\bmw_i^{(\ell)}$. Let $\bmv_j$ and $\bmv'_j$ be the two normal vectors of the $j$th \intersectionpoint. Then, the correct forward matrix up to layer $\ell-1$ is given by
$\bfF^{(\ell-1)}_{j} \allowbreak = \allowbreak \bfD^{(\ell-1)}_j \allowbreak \bfW^{(\ell-1)} \allowbreak \bfD^{(\ell-2)}_j \allowbreak \bfW^{(\ell-2)} \cdots \bfD^{(1)}_j \bfW^{(1)}$.
Here, we just write $\bfF^{(\ell-1)}_{j}$ instead of $\bfF^{(\ell-1)}_{\bms_j}$ for simplicity. Since $\bfN_j$ denotes a matrix whose columns form an orthonormal basis of the orthogonal complement $\Span\{\bmv_j, \bmv'_j\}^\bot$, we have $(\bfF^{(\ell-1)}_j \bfN_j)^\top \bmw_i^{(\ell)} = \bm{0}$. Equivalently,
$\bmw_i^{(\ell)\top} (\bfF^{(\ell-1)}_j \bfN_j)\allowbreak (\bfF^{(\ell-1)}_j \bfN_j)^\top \allowbreak \bmw_i^{(\ell)} \allowbreak = 0$.
Since this must hold for all $j \in [m]$, it follows that
$\bmw_i^{(\ell)\top} \sum_j (\bfF^{(\ell-1)}_j \bfN_j)\allowbreak (\bfF^{(\ell-1)}_j \bfN_j)^\top \bmw_i^{(\ell)} = 0$.
Therefore, when $m$ is sufficiently large, the eigenvector corresponding to the minimum eigenvalue (zero) of $\sum_j (\bfF^{(\ell-1)}_j \bfN_j)(\bfF^{(\ell-1)}_j \bfN_j)^\top$ coincides with the neuron's signature. This forms the basis of neuron extraction.  

Now, suppose that the $k$th neuron in layer $\ell-1$ is persistent and cannot be recovered. In this case, we approximate its weight and bias as zero, which is equivalent to setting the corresponding diagonal entry of the activation matrix to zero. Thus, the modified diagonal matrix is given by $\tilde\bfD^{(\ell-1)}_j = \bfD^{(\ell-1)}_j - \bme_k \bme_k^\top$. Accordingly, the incorrect forward matrix becomes
\begin{align}
    \tilde\bfF^{(\ell-1)}_j &= \tilde\bfD^{(\ell-1)}_j \bfW^{(\ell-1)} \bfD^{(\ell-2)}_j \bfW^{(\ell-2)} \cdots  \bfD^{(1)}_j \bfW^{(1)} \\
    &= \bfF^{(\ell-1)}_j - \bme_k \bme_k^\top \bfW^{(\ell-1)} \bfD^{(\ell-2)}_j \cdots \bfD^{(1)}_j \bfW^{(1)}.
\end{align}
To simplify notation, for each index $j$, we denote the output corresponding to the $k$-th neuron at layer $\ell-1$ after applying the local linear map to $\mathbf{N}_j$ by $\bm{z}_{k, j}^{(\ell-1)\top} := \bm{w}_k^{(\ell-1)\top}\,\mathbf{F}_j^{(\ell-2)}\,\mathbf{N}_j$. Since $\mathbf{N}_j$ is a matrix consisting of $d_0-2$ orthonormal basis vectors, the vector $\bm{z}_{k,j}^{(\ell-1)}$ has $d_0-2$ entries. More generally, let $\bm{z}_{r,j}^{(\ell-1)\top}$ denote the $r$th row of $\mathbf{F}_j^{(\ell-1)}\mathbf{N}_j$. Using this notation, after multiplying by $\mathbf N_j$, we can rewrite $\tilde{\mathbf{F}}_j^{(\ell-1)}\mathbf{N}_j = \mathbf{F}_j^{(\ell-1)}\mathbf{N}_j - \bm{e}_k\,\bm{z}_{k,j}^{(\ell-1)\top}$, where $\bm{e}_k$ is the $k$-th standard basis vector. In addition, the correct local linear mapping can be expressed as $\mathbf{F}_j^{(\ell-1)} \mathbf{N}_j = \bigl( \bm{z}_{1, j}^{(\ell-1)}, \bm{z}_{2, j}^{(\ell-1)}, \dots, \bm{z}_{d_{\ell-1}, j}^{(\ell-1)} \bigr)^\top$. When a persistent neuron exists at index $k$, the (erroneous) local linear mapping becomes
$\tilde{\mathbf{F}}_j^{(\ell-1)} \mathbf{N}_j = \bigl( \bm{z}_{1, j}^{(\ell-1)}, \dots, \bm{z}_{k-1, j}^{(\ell-1)}, \mathbf{0}, \bm{z}_{k+1, j}^{(\ell-1)}, \dots, \bm{z}_{d_{\ell-1}, j}^{(\ell-1)} \bigr)^\top$.

An attacker uses this erroneous matrix to estimate the signature by solving the following minimization problem for the corresponding quadratic form:
\begin{equation}
\min_{\substack{\|\bm{w}\| = 1 \\ (\bm{w})_k = 0}}\ \bm{w}^\top \sum_j
\Bigl(
(\tilde{\mathbf{F}}_j^{(\ell-1)} \mathbf{N}_j)\,
(\tilde{\mathbf{F}}_j^{(\ell-1)} \mathbf{N}_j)^\top
\Bigr)
\bm{w},
\label{eq:attacker_obj}
\end{equation}
where $(\bm{w})_k$ is the $k$th coordinate of vector $\bm{w}$. The constraint $(\bm{w})_k=0$ excludes the trivial missing-coordinate direction. Indeed, if the quadratic form were minimized over all unit vectors in $\mathbb{R}^{d_{\ell-1}}$, the vector $\bm e_k$ would attain value zero, because the $k$th row and column are zero. This vector is not a candidate signature, but only represents the coordinate of the unrecovered persistent neuron.

Since the $k$-th row of $\tilde{\mathbf{F}}_j^{(\ell-1)} \mathbf{N}_j$ is zero for all $j$, \cref{eq:attacker_obj} depends only on the vector $\bm w_{\setminus k}$ obtained by removing the $k$-th coordinate of $\bm w$. Therefore, the projected problem is equivalently written as
\begin{equation}
\min_{\substack{\|\bm{w}_{\setminus k}\| = 1}}
\ \bm{w}_{\setminus k}^\top
\sum_j
\mathbf{F}_{j,\setminus k}^{(\ell-1)} \mathbf{N}_j
\left(
\mathbf{F}_{j,\setminus k}^{(\ell-1)} \mathbf{N}_j
\right)^\top
\bm{w}_{\setminus k},
\label{eq:reduced_signature_obj_thm}
\end{equation}
where $\mathbf{F}_{j,\setminus k}^{(\ell-1)}$ is obtained by removing the $k$-th row of $\mathbf{F}_{j}^{(\ell-1)}$. If a persistent neuron $\bmw_{k}^{(\ell-1)}$ exists in the previous layer, then its weight and bias cannot be identified by the attacker and are therefore effectively treated as zero. In this case, estimating the signature of a neuron in the next layer is equivalent to using the local linear mapping with the corresponding row removed, namely $\bfF_{j,\setminus k}^{(\ell-1)}$. Accordingly, the estimated signature also excludes the corresponding coordinate, and the optimization is carried out over a $(d_{\ell-1}-1)$-dimensional vector $\bmw_{\setminus k}$ instead of the original full-dimensional signature vector. Consequently, the attacker's objective is no longer to recover the complete signature $\bmw_{i}^{(\ell)}$, but rather the reduced signature $\bmw_{i,\setminus k}^{(\ell)}$ with the $k$-th component removed.

\subsubsection{Discrepancy from the Exact Signature Estimation.}

By the reduction above, when a persistent neuron exists, signature recovery is equivalent to estimating the signature using the matrix obtained by removing the $k$-th row from the local linear mapping. Since this matrix differs from the true local linear mapping in the presence of a persistent neuron, substituting the true reduced signature $\bm{w}_{i,\setminus k}^{(\ell)}$ into \cref{eq:reduced_signature_obj_thm} does not, in general, yield zero. Consequently, minimizing \cref{eq:reduced_signature_obj_thm} does not necessarily recover $\bm{w}_{i,\setminus k}^{(\ell)}$.

To quantify this discrepancy, we next derive a bound on the angle between the true reduced signature and the signature obtained from \cref{eq:reduced_signature_obj_thm}. The following theorem provides an explicit upper bound on $\tan \theta$, where $\theta$ denotes the angle between the true signature and the signature estimated by \cref{eq:reduced_signature_obj_thm}.

\begin{theorem}[Upper Bound on Signature Discrepancy] \label{thm:upper_boud_tan} \\
Let $\tilde{\bfG} = \frac{1}{m} \sum_j \mathbf{F}_{j,\setminus k}^{(\ell-1)} \mathbf{N}_j \left( \mathbf{F}_{j,\setminus k}^{(\ell-1)} \mathbf{N}_j \right)^\top$ be the matrix used in the minimization problem \cref{eq:reduced_signature_obj_thm}, where the normalization by $m$ is introduced only for convenience and does not change the solution of \cref{eq:reduced_signature_obj_thm}. Define
$\bfG = \tilde{\bfG} - \bm{\beta}\bm{\beta}^\top/\alpha$,
where $\alpha = \frac{1}{m} \sum_j \bm{z}_{k,j}^{(\ell-1)\top} \bm{z}_{k,j}^{(\ell-1)}$, and
\begin{multline}
\bm{\beta} = \frac{1}{m} \sum_j \mathbf{F}_{j,\setminus k}^{(\ell-1)} \mathbf{N}_j \, \bm{z}_{k,j}^{(\ell-1)}
= \frac{1}{m} \sum_j
(
\bm{z}_{1,j}^{(\ell-1)\top} \bm{z}_{k,j}^{(\ell-1)},
\bm{z}_{2,j}^{(\ell-1)\top} \bm{z}_{k,j}^{(\ell-1)},
\dots, \\
\bm{z}_{k-1,j}^{(\ell-1)\top} \bm{z}_{k,j}^{(\ell-1)},
\bm{z}_{k+1,j}^{(\ell-1)\top} \bm{z}_{k,j}^{(\ell-1)},
\dots,
\bm{z}_{d_{\ell-1},j}^{(\ell-1)\top} \bm{z}_{k,j}^{(\ell-1)}
)^\top.
\end{multline}
Assume that $\tilde{\bfG} \in \mathbb{R}^{(d_{\ell-1}-1)\times(d_{\ell-1}-1)}$ has full rank $d_{\ell-1}-1$, and that $\bm{\beta}$ is not orthogonal to any eigenvector of $\tilde{\bfG}$. Consider the eigendecomposition $\bfG = \sum_\eta \mu_\eta \bmq_\eta \bmq_\eta^\top$, where the eigenvalues satisfy $0 = \mu_1 < \mu_2 \leq \cdots \leq \mu_{d_{\ell-1}-1}$. Then the eigenvector $\bmq_1$ coincides with the true reduced signature $\bmw_{i,\setminus k}^{(\ell)}$. In addition, let $\theta$ be the angle between the signature estimated via \cref{eq:reduced_signature_obj_thm} and the true reduced signature. Then
\begin{align}
\tan \theta
\le
\frac{
|\bm{w}_{i,\setminus k}^{(\ell)\top}\bm{\beta}|
}{
\alpha + \bm{\beta}^\top \bfG^\dagger \bm{\beta}
}
\sqrt{
\sum_{\eta=2}
\frac{(\bm{q}_\eta^\top \bm{\beta})^2}{(\mu_\eta - \lambda_1)^2}
},
\label{eq:tan_upper_thm}
\end{align}
where $\lambda_1$ is the smallest eigenvalue of $\tilde{\bfG}$ and
$\bfG^\dagger = \sum_{\eta=2} \bmq_\eta \bmq_\eta^\top/\mu_\eta$ denotes the pseudoinverse of $\bfG$.
\end{theorem}
Before proving the theorem, we comment on the assumptions used above. First, we assume that $\tilde{\bfG} \in \mathbb{R}^{(d_{\ell-1}-1)\times(d_{\ell-1}-1)}$ has full rank $d_{\ell-1}-1$. This is the same generic rank condition used in layer-wise extraction: writing $\mathbf A_j=\mathbf F_{j,\setminus k}^{(\ell-1)}\mathbf N_j$, it is equivalent to requiring the stacked matrix $[\mathbf A_1,\ldots,\mathbf A_m]$ to have full row rank. Although each individual $\mathbf A_j$ may be low rank, different \intersectionpoints induce different local subspaces, and their concatenation can span the full reduced layer as $m$ increases. We also confirmed this behavior in a simple expanding-network experiment on the Iris dataset, using an MLP with input dimension $4$, hidden width $10$, and output dimension $3$, where $\tilde{\bfG}$ reached full rank after collecting sufficiently many \intersectionpoints. In our analysis, we assume that sufficiently many \intersectionpoints are collected so that this condition holds.

We also assume that the vector $\bm{\beta}$ is not orthogonal to any eigenvector of $\tilde{\bfG}$. Since neural networks are trained in a stochastic manner and these quantities depend on many continuous parameters, exact orthogonality is highly unlikely in practice, and its probability is expected to be negligible.

\begin{proof}
First, we prove that the eigenvector corresponding to the smallest eigenvalue of $\bfG$ coincides with the true reduced signature $\bmw_{i,\setminus k}^{(\ell)}$. We then derive the stated upper bound on $\tan \theta$.

\paragraph{(1) Proof that $\bmq_1 = \bmw_{i,\setminus k}^{(\ell)}$.}
Consider the matrix $\frac{1}{m} \sum_j \mathbf{F}_{j}^{(\ell-1)} \mathbf{N}_j \left( \mathbf{F}_{j}^{(\ell-1)} \mathbf{N}_{j} \right)^\top$, and permute its rows and columns so that the $k$-th row and column are moved to the first row and column. Denote the resulting matrix by
%\begin{align}
$\tilde{\mathbf{G}}_{\mathrm{mov}}
=
\begin{pmatrix}
\alpha & \bm{\beta}^\top \\
\bm{\beta} & \tilde{\mathbf{G}}
\end{pmatrix}$.
%\end{align}
For this matrix, a zero-eigenvalue eigenvector is given by $(w_{i,k}^{(\ell)}, \bm{w}_{i,\setminus k}^{(\ell)\top})^\top$, where $w_{i,k}^{(\ell)}$ denotes the $k$-th component of the signature of the $i$-th neuron at layer $\ell$. Hence,
$\tilde{\mathbf{G}}_{\mathrm{mov}} (w_{i,k}^{(\ell)}, \bm{w}_{i,\setminus k}^{(\ell)\top})^\top = 0$,
which implies
\begin{align}
\left\{
\begin{aligned}
& \alpha\, w_{i,k}^{(\ell)} + \bm{\beta}^\top \bm{w}_{i,\setminus k}^{(\ell)} = 0, \\
& \bm{\beta}\, w_{i,k}^{(\ell)} + \tilde{\mathbf{G}}\, \bm{w}_{i,\setminus k}^{(\ell)} = 0.
\end{aligned}
\right.
\end{align}
Solving the first equation gives $w_{i,k}^{(\ell)} = - \bm{\beta}^\top \bm{w}_{i,\setminus k}^{(\ell)} / \alpha$. Substituting this into the second equation yields
$\left(\tilde{\mathbf{G}} - \bm{\beta}\bm{\beta}^\top/\alpha \right)
\bm{w}_{i,\setminus k}^{(\ell)} = 0$.
Thus, by definition of $\bfG$, the vector $\bm{w}_{i,\setminus k}^{(\ell)}$ is the eigenvector of $\bfG$ associated with eigenvalue zero.

%\paragraph{(2) Proof that $\bmp \propto (\bfG - \lambda \bfI)^{-1} \bm{\beta}$.}
\paragraph{(2) Representation of eigenvalues of $\tilde{\bfG}$.}
To derive the bound on $\tan\theta$, consider an eigenpair $(\lambda,\bmp)$ of $\tilde{\bfG}$ with $\lambda$ not an eigenvalue of $\bfG$. Then from $\tilde{\bfG}\bmp = \lambda \bmp$ and $\tilde{\bfG} = \bfG + \bm{\beta}\bm{\beta}^\top/\alpha$, we obtain $(\bfG - \lambda \bfI)\bmp = -(\bm{\beta}^\top \bmp)\bm{\beta}/\alpha$. Since $\bfG - \lambda \bfI$ is invertible, this gives
$\bmp = -(\bm{\beta}^\top \bmp)(\bfG - \lambda \bfI)^{-1}\bm{\beta}/\alpha$,
and hence $\bmp \propto (\bfG - \lambda \bfI)^{-1}\bm{\beta}$,
where $\propto$ denotes equality up to a nonzero scalar factor. Left-multiplying by $\bm{\beta}^\top$ and dividing by $\bm{\beta}^\top \bmp$ yields
\begin{align}
1 + \bm{\beta}^\top (\bfG - \lambda \bfI)^{-1} \bm{\beta}/\alpha = 0.
\label{eq:eig}
\end{align}

\paragraph{(3) Upper bound on $\lambda_1$.}
Using the eigendecomposition of $\bfG$, we obtain
$\bm{\beta}^\top (\bfG - \lambda \bfI)^{-1} \bm{\beta} = -\frac{(\bm{w}_{i,\setminus k}^{(\ell)\top}\bm{\beta})^2}{\lambda_1} + \sum_{\eta=2} \frac{(\bm{q}_\eta^\top \bm{\beta})^2}{\mu_\eta - \lambda_1}$. Combining with \cref{eq:eig} gives $\alpha - \frac{(\bm{w}_{i,\setminus k}^{(\ell)\top}\bm{\beta})^2}{\lambda_1} + \sum_{\eta=2} \frac{(\bm{q}_\eta^\top \bm{\beta})^2}{\mu_\eta - \lambda_1} = 0$. Let $\bfG^\dagger = \sum_{\eta=2} \bmq_\eta \bmq_\eta^\top/\mu_\eta$. Since the last sum is bounded below by replacing $\mu_\eta-\lambda_1$ with $\mu_\eta$, we obtain
\begin{align}
\lambda_1
\le
\frac{(\bm{w}_{i,\setminus k}^{(\ell)\top}\bm{\beta})^2}
{\alpha + \bm{\beta}^\top \bfG^\dagger \bm{\beta}}.
\label{eq:lambda_upper}
\end{align}

\paragraph{(4) Upper bound on $\tan \theta$.}
From step (2), the eigenvector associated with $\lambda_1$ can be written as
$\hat{\bmp}_1 = -\frac{\bm{w}_{i,\setminus k}^{(\ell)\top}\bm{\beta}}{\lambda_1} \bm{w}_{i,\setminus k}^{(\ell)} + \sum_{\eta=2} \frac{\bm{q}_\eta^\top \bm{\beta}}{\mu_\eta - \lambda_1} \bm{q}_\eta$.
From this decomposition,
$\tan \theta
=
\left\|
\sum_{\eta=2}
\frac{\bm{q}_\eta^\top \bm{\beta}}{\mu_\eta - \lambda_1}
\bm{q}_\eta
\right\| \Big/
\left|
\frac{\bm{w}_{i,\setminus k}^{(\ell)\top}\bm{\beta}}{\lambda_1}
\right|
$.
Substituting \cref{eq:lambda_upper} yields \cref{eq:tan_upper_thm}.
\qed
\end{proof}

\subsubsection{Discussion on \cref{thm:upper_boud_tan}.}

Based on \cref{thm:upper_boud_tan}, we explain why the estimation error is difficult to avoid within the existing method. The right-hand side of \cref{eq:tan_upper_thm} is the product of the inner-product factor $|\bmw_{i,\setminus k}^{(\ell)\top}\bm{\beta}|$ and the factor $(\alpha+\bm{\beta}^\top\bfG^\dagger\bm{\beta})^{-1}\sqrt{\sum_{\eta=2}(\bmq_\eta^\top\bm{\beta})^2/(\mu_\eta-\lambda_1)^2}$. Thus, if $|\bmw_{i,\setminus k}^{(\ell)\top}\bm{\beta}| = 0$, then the bound gives $\tan\theta = 0$. The second factor can also vanish, but this requires $\bm{\beta}$ to have no component in the directions $\bmq_\eta$ for $\eta \geq 2$, i.e., $\bm{\beta}$ must be aligned with the true reduced signature $\bmw_{i, \setminus k}^{(\ell)}$. These are special alignments, and we do not expect them to occur in general. Since \cref{eq:tan_upper_thm} provides only an upper bound on the error, these conditions are sufficient but not necessarily necessary. However, as shown empirically in \cref{subsec:error_experiment}, the bound closely matches the observed values of $\tan\theta$, which indicates that these factors explain the observed estimation error.

The role of the inner-product factor can also be understood from the fact, shown in \cref{thm:upper_boud_tan}, that the matrix that yields the true reduced signature is not $\tilde{\mathbf{G}} = \frac{1}{m} \sum_j \bfF_{j,\setminus k}^{(\ell-1)} \bfN_j (\bfF_{j,\setminus k}^{(\ell-1)} \bfN_j)^\top$, but rather $\bfG = \tilde{\mathbf{G}} - \bm{\beta}\bm{\beta}^\top / \alpha$. If the true reduced signature $\bm{w}_{i,\setminus k}^{(\ell)}$ is orthogonal to $\boldsymbol{\beta}$, then $\tilde{\mathbf{G}}\,\bm{w}_{i,\setminus k}^{(\ell)} = (\bfG + \boldsymbol{\beta}\boldsymbol{\beta}^\top/\alpha)\bm{w}_{i,\setminus k}^{(\ell)} = 0$, and hence $\tilde{\mathbf{G}}$ still has $\bm{w}_{i,\setminus k}^{(\ell)}$ as an eigenvector associated with eigenvalue $0$. In this case, the signature can still be recovered correctly.

However, it is unlikely that these vectors are exactly orthogonal. Indeed, since $\boldsymbol{\beta} = \frac{1}{m} \sum_j \bfF_{j,\setminus k}^{(\ell-1)} \bfN_j\,\bmz_{k,j}^{(\ell-1)}$, taking the inner product with the true reduced signature gives $\bm{w}_{i,\setminus k}^{(\ell)\top}\boldsymbol{\beta} = \frac{1}{m} \sum_j \bm{w}_{i,\setminus k}^{(\ell)\top} \bfF_{j,\setminus k}^{(\ell-1)} \bfN_j\, \bmz_{k,j}^{(\ell-1)}$. Here, $\tilde{\bm{z}}_{i,j}^{(\ell)} = \bm{w}_{i,\setminus k}^{(\ell)\top} \bfF_{j,\setminus k}^{(\ell-1)} \bfN_j$ can be interpreted as the local variation of the pre-activation of the $i$-th neuron at layer $\ell$ computed while ignoring the persistent neuron. Therefore, the above inner product is essentially an average of products between these local variations and the local variations of the $k$-th neuron in layer $\ell-1$. In a trained network, there is no reason to expect these local variations to be exactly orthogonal on average. Similarly, there is no reason to expect $\boldsymbol{\beta}$ to be exactly aligned with the true reduced signature. Thus, the exceptional cases where the bound vanishes are unlikely in practice.

%Therefore, when a persistent neuron exists, an error term is expected to be introduced into the signature estimation. The magnitude of this error is evaluated experimentally in the next subsection.

From these considerations, when a persistent neuron exists, an unavoidable error term is expected to be introduced into the signature estimation. The magnitude of this error is evaluated experimentally in the next subsection.

\subsection{Experimental Results of Estimated Errors} \label{subsec:error_experiment}

\begin{figure}[t]
    \centering
    \begin{subfigure}[t]{0.45\textwidth}
        \centering
        \includegraphics[width=\linewidth]{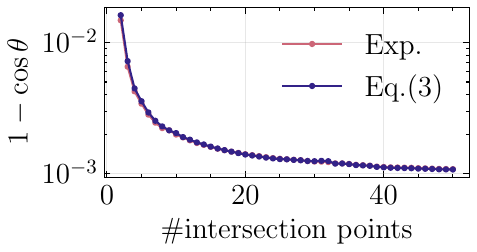}
        \caption{MNIST.}
        \label{subfig:prob_MNIST_est_error_int_points}
    \end{subfigure}
    \begin{subfigure}[t]{0.45\textwidth}
        \centering
        \includegraphics[width=\linewidth]{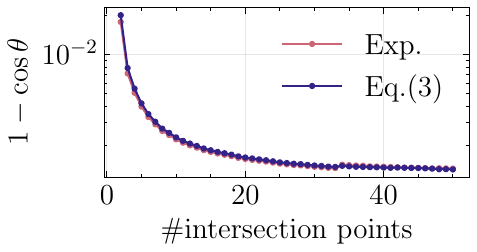}
        \caption{FMNIST.}
        \label{subfig:prob_FMNIST_est_error_int_points}
    \end{subfigure}
    \caption{Geometric average of $1-\cos \theta$ between the true and estimated signatures for models with a persistent neuron at width 256.}
    \label{fig:est_error_int_points}
\end{figure}

\begin{figure}[t]
    \centering
    \begin{subfigure}[t]{0.45\textwidth}
        \centering
        \includegraphics[width=\linewidth]{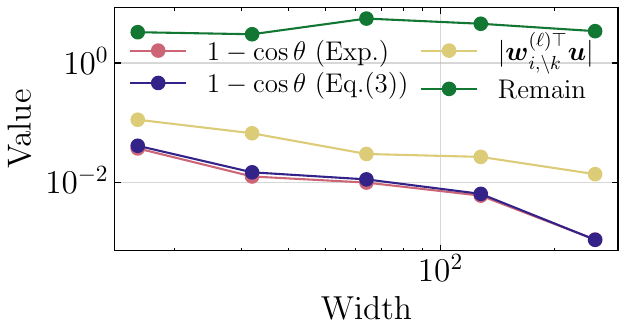}
        \caption{MNIST.}
        \label{subfig:prob_MNIST_est_error}
    \end{subfigure}
    \begin{subfigure}[t]{0.45\textwidth}
        \centering
        \includegraphics[width=\linewidth]{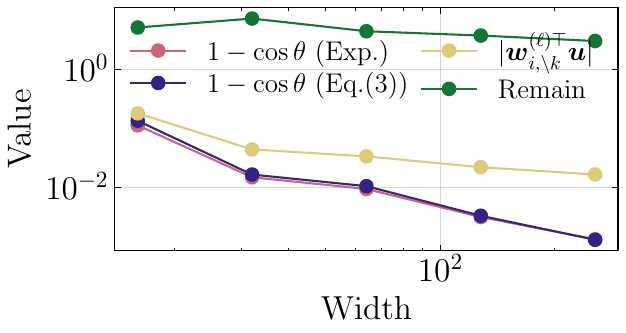}
        \caption{FMNIST.}
        \label{subfig:prob_FMNIST_est_error}
    \end{subfigure}
    \caption{Geometric average of $1-\cos \theta$ between true and estimated signatures with a persistent neuron while varying the model width.}
    \label{fig:est_error}
\end{figure}

We next quantify the noise introduced into neuron estimation by persistent neurons. We show experimentally that the estimated signature contains noise between $O(1/d)$ and $O(1/(d\sqrt{d}))$, where $d$ is the model width, and interpret this scaling using \cref{eq:tan_upper_thm}.

\subsubsection{Experimental Results.}

The model training conditions are essentially the same as those in \cref{subsec:experiment_queries}, except that we used five different widths: 16, 32, 64, 128, and 256. For each trained model, we considered the case where one persistent neuron in the 9th layer could not be recovered, and we then attempted to recover neurons in the 10th layer.

%For recovery in the 10th layer, we extracted $10^7$ \intersectionpoints for each model and used only those neurons in the 10th layer for which at least 50 \intersectionpoints were extracted. For these neurons, weight recovery used 50 \intersectionpoints. Since using only 50 \intersectionpoints does not necessarily allow all dimensions of each neuron's signature to be obtained (depending on the activation patterns of the previous layer), we computed the angle $\theta$ from the true signature only for the components that were successfully estimated.

For 10th-layer recovery, we extracted $10^7$ \intersectionpoints per model and retained only neurons with at least 50 \intersectionpoints. We recovered each such weight using 50 \intersectionpoints. Since 50 \intersectionpoints may not recover all signature dimensions (depending on previous-layer activation patterns), we computed the angle $\theta$ to the true signature only on the successfully estimated components.

%\cref{fig:est_error_int_points} shows the $1-\cos \theta$ when varying the number of \intersectionpoints used for estimating the target neuron of a trained model with width 256. In the figure, ``Exp.'' denotes the empirical results obtained when neuron recovery is performed in the presence of a persistent neuron, while ``Eq.~(3)'' denotes the upper bound on $1-\cos\theta$ computed from \cref{eq:tan_upper_thm}. Since \cref{eq:tan_upper_thm} provides an upper bound on $\tan\theta$, we convert it into an upper bound on $1-\cos\theta$ using $\cos\theta = 1/\sqrt{1+\tan^2\theta}$. As seen from the figure, our upper bound closely matches the experimental values and explains the observed behavior well. We can observe that the estimation error of the neuron signature decreases as the number of \intersectionpoints increases, but the amount of decrease is gradually reduced. Therefore, it is difficult in practice to eliminate the estimation error solely by increasing the number of \intersectionpoints.

\cref{fig:est_error_int_points} shows $1-\cos \theta$ as the number of \intersectionpoints varies for a trained width-256 model. ``Exp.'' denotes empirical recovery with a persistent neuron, and ``Eq.~(3)'' denotes the bound on $1-\cos\theta$ obtained from \cref{eq:tan_upper_thm} via $\cos\theta = 1/\sqrt{1+\tan^2\theta}$. The bound closely matches the experiments. Although the error decreases with more \intersectionpoints, the decrease gradually saturates, making it difficult to eliminate the error by increasing \intersectionpoints alone.

\cref{fig:est_error} shows the results for $1-\cos\theta$ as a function of the model width. For each width, we used the number of \intersectionpoints that achieved the smallest empirical value of $1-\cos\theta$. To clarify which terms dominate the bound, we plot the following quantities. By normalizing the numerator and denominator of \cref{eq:tan_upper_thm} using $\|\boldsymbol{\beta}\|$, the bound can be rewritten as
\begin{align}
    \tan \theta
    \le
    \frac{
        |\bm{w}_{i,\setminus k}^{(\ell)\top}\bm{u}|
    }{
        \alpha/\|\bm{\beta}\|^2 + \bm{u}^\top \bfG^\dagger \bm{u}
    }
    \sqrt{
        \sum_{\eta=2}
        \frac{(\bm{q}_\eta^\top \bm{u})^2}{(\mu_\eta - \lambda_1)^2}
    }, \label{eq:tan_upper_u}
\end{align}
%where $\bm{u} = \bm{\beta}/\|\bm{\beta}\|$. Among the terms on the right-hand side, we separately plot $|\bm{w}_{i,\setminus k}^{(\ell)\top}\bm{u}|$ and the remaining factor $(\alpha/\|\bm{\beta}\|^2 + \bm{u}^\top \bfG^\dagger \bm{u})^{-1} \sqrt{\sum_{\eta=2} \frac{(\bm{q}_\eta^\top \bm{u})^2}{(\mu_\eta - \lambda_1)^2}}$. The label ``Remain'' in the figure refers to this latter factor. When $\theta$ is sufficiently small, the approximation $1-\cos\theta \approx \frac{1}{2}\tan^2\theta$ holds, and thus $1-\cos\theta$ is approximately determined by the product of these two terms. More precisely, since both axes use a log scale, the slope of $1-\cos\theta$ corresponds to twice the sum of the slopes of these two factors. As shown in the figure, the experimentally measured $1-\cos\theta$ closely matches the value predicted from \cref{eq:tan_upper_u}.
where $\bm{u} = \bm{\beta}/\|\bm{\beta}\|$. We separately plot $|\bm{w}_{i,\setminus k}^{(\ell)\top}\bm{u}|$. The curve labeled ``Remain'' is $(\alpha/\|\bm{\beta}\|^2 + \bm{u}^\top \bfG^\dagger \bm{u})^{-1}\allowbreak \sqrt{\sum_{\eta=2} \frac{(\bm{q}_\eta^\top \bm{u})^2}{(\mu_\eta - \lambda_1)^2}}$. For small $\theta$, $1-\cos\theta \approx \tan^2\theta/2$, so $1-\cos\theta$ is approximately determined by the product of these two factors; on log-log axes, its slope is twice the sum of their slopes. The measured $1-\cos\theta$ closely matches the value predicted by \cref{eq:tan_upper_u}.

From the figure, we observe that the empirical values of $1-\cos\theta$ decrease approximately linearly in the log–log plot. This indicates a power-law dependence on the model width $d$, and we therefore perform a linear fit in the log–log scale. The resulting slopes are $-1.12$ for MNIST and $-1.51$ for FMNIST. These values suggest that, for width $d$, the estimation error decreases at a rate between $O(1/d)$ and $O(1/(d\sqrt{d}))$. The source of this improvement can also be inferred from the figure: since the ``Remain'' term is nearly flat across widths, the dominant contribution comes from the decay of $|\bm{w}_{i,\setminus k}^{(\ell)\top}\bm{u}|$.

From these experimental results, we conclude that the estimation accuracy saturates even if the number of \intersectionpoints is further increased, and that the saturation level scales approximately as $O(1/d)$ to $O(1/(d\sqrt{d}))$ with respect to the model width $d$.

\subsubsection{Discussion.}

%Here, we analyze the estimation error of the signature from two perspectives: the number of \intersectionpoints and the model width, and argue that removing this error is difficult in practice. We first consider the effect of the number of \intersectionpoints. As shown in \cref{fig:est_error_int_points}, the value of $1-\cos\theta$ decreases monotonically as the number of \intersectionpoints increases. From this experimental result alone, it may appear that the error can be eliminated simply by increasing the number of \intersectionpoints.
%
%To examine the rate at which the error decreases, we take logarithms of both $1-\cos\theta$ and the number of \intersectionpoints, and perform a linear regression between them. The resulting slopes are $-0.59$ for MNIST and $-0.58$ for FMNIST. Approximating these values by $-1/2$, this indicates that when the number of \intersectionpoints is $m$, the quantity $1-\cos\theta$ decays as $O(1/\sqrt{m})$. Intuitively, this behavior is consistent with an error reduction governed by a central limit type effect. On the other hand, this rate implies that improving the error by one order of magnitude requires roughly 100 times more \intersectionpoints. In other words, achieving each additional digit of accuracy requires exponentially more \intersectionpoints, suggesting that eliminating the error by increasing the number of \intersectionpoints is not feasible in practice.

Here, we discuss why the estimation accuracy improves at approximately a rate between $O(1/d)$ and $O(1/(d\sqrt{d}))$ with respect to the width $d$, based on \cref{eq:tan_upper_u}. Since a precise analysis for fully trained models is difficult, we instead consider a highly simplified setting. Specifically, we assume that the vector $\boldsymbol{\beta}$ (and hence $\boldsymbol{u}$) is independent of $\bm{w}_{i,\setminus k}^{(\ell)}$, and that $\bm{w}_{i,\setminus k}^{(\ell)}$ is uniformly distributed on the unit sphere. Although this assumption does not necessarily hold for trained models, at least under Kaiming normal initialization each weight is sampled from a Gaussian distribution. Therefore, it is reasonable to assume that the normalized weight vector $\bm{w}_{i,\setminus k}^{(\ell)}$ is uniformly distributed on the unit hypersphere at initialization.

Under this assumption, when $d$ is sufficiently large, we have approximately $|\bm{w}_{i,\setminus k}^{(\ell)\top}\bm{u}| = O(1/\sqrt{d})$. We have experimentally confirmed that the remaining factor $(\alpha/\|\boldsymbol{\beta}\|^2 + \boldsymbol{u}^\top \bfG^\dagger \boldsymbol{u})^{-1} \sqrt{\sum_{\eta=2} \frac{(\boldsymbol{q}_\eta^\top \boldsymbol{u})^2}{(\mu_\eta - \lambda_1)^2}}$ is nearly independent of the width. This implies that $\tan\theta$ decreases approximately as $O(1/\sqrt{d})$ as the width increases. When $\theta$ is sufficiently small, the approximation $1 - \cos\theta \approx \frac{1}{2}\tan^2\theta$ holds, which suggests that $1 - \cos\theta = O(1/d)$. This provides a rough explanation for why the estimation accuracy improves at a rate between $O(1/d)$ and $O(1/(d\sqrt{d}))$. Of course, in practice an attacker cannot control or increase the width of the victim model's neural network architecture. Hence, reducing the error through width scaling is not a viable option for the attacker. From this perspective as well, removing the estimation error is inherently difficult.

%%%%%%%%%%%%%%%%%%%%%%%%%%%%%%%%%%%%%%%%%%%%%%%%%%%
%%%%%%%%%%%%%%%%%%%%%%%%%%%%%%%%%%%%%%%%%%%%%%%%%%%
\section{Cross-Layer Extraction for Persistent Neurons} \label{sec:bypass}
As discussed above, the presence of persistent neurons prevents their weight vectors from being recovered by existing techniques. An attacker cannot observe the corresponding activation boundaries and therefore cannot extract the associated weights. Unlike dead neurons, persistent neurons affect subsequent layers, which makes recovering their contributions necessary for model extraction.
Since \intersectionpoints for persistent neurons are not available, we need an alternative method. We recover their contribution via a cross-layer boundary, exploiting decision and activation boundaries of neurons in a deeper layer. 

Let $\mathcal{P}$ be the index set of persistent neurons in the $\ell$th layer. Existing model extraction fails to recover the weight $\bm w^{(\ell)}_j$ for $j \in \mathcal{P}$. On the other hand, when a point lies on the activation boundary at depth $\ell + 1$ and neuron index $k$, we still have the following relation:
\begin{align}
    - b_k^{(\ell + 1)} 
    &= \bm w^{(\ell + 1)\top}_{k} \bm z^{(\ell)} 
    = \bm w^{(\ell + 1)\top}_{k,\bar{\mathcal{P}}} \bm z^{(\ell)}_{\bar{\mathcal{P}}} + \sum_{j \in \mathcal{P}} w^{(\ell+1)}_{k,j} (\bm w^{(\ell)\top}_{j} \bm z^{(\ell-1)} + b_j^{(\ell)}),   \\
    &= \bm w^{(\ell + 1)\top}_{k,\bar{\mathcal{P}}} \bm z^{(\ell)}_{\bar{\mathcal{P}}} + \left( \sum_{j \in \mathcal{P}} w^{(\ell + 1)}_{k,j} \bm w^{(\ell)}_{j} \right)^\top \bm z^{(\ell-1)} + \sum_{j \in \mathcal{P}} w^{(\ell+1)}_{k,j} b_j^{(\ell)},   
\end{align}
We denote by $\bm a_\mathcal{P}$ the subvector of $\bm a$ consisting of the components indexed by $\mathcal{P}$, and by $\bm a_{\bar{\mathcal{P}}}$ the subvector consisting of the remaining components.
Define 
\[ \bm w_{cross} = \begin{pmatrix}\bm w_{k,\bar{\mathcal{P}}}^{(\ell+1)} \\ \sum_{j \in \mathcal{P}} w^{(\ell+1)}_{k,j} \bm w^{(\ell)}_{j} \end{pmatrix}. \]
Then, points on the activation boundary satisfy an affine constraint of the form $\bm w_{cross}^\top \begin{pmatrix} \bm z^{(\ell)}_{\bar{\mathcal{P}}} \\  \bm z^{(\ell - 1)} \end{pmatrix} = \mathrm{const}$. Equivalently, for any two boundary points, their difference is orthogonal to $\bm w_{cross}$. 
We collect multiple points on the boundary, take differences against a fixed reference point, and recover $\bm w_{cross}$ up to scale as a normal vector of the subspace spanned by these differences. 
This vector contains a linear combination of the persistent neuron weights $\bm w_j^{(\ell)}$ for $j \in \mathcal{P}$. 

Some important considerations must be made regarding cross-layer extraction. 
First, it is necessary to modify the corresponding consistency algorithm. While the existing algorithm analyzes each \intersectionspace pairwise, cross-layer extraction requires analyzing multiple \intersectionspaces simultaneously. 
Second, cross-layer analysis can extract only the span of the original weight vectors of persistent neurons. More precisely, since persistent neurons are always active and are never suppressed by ReLU, their corresponding signs and biases are linearly combined into the next layer. Unless they become inactive, we cannot extract them individually. Thus, in the recovered model, we need exceptional treatment for them. Specifically, any basis of the span can be used, and the corresponding ReLU must be removed.

\subsection{Modified Consistency Algorithm}
Let $\bm s \in \mathbb{R}^{d_0}$ be an \intersectionpoint, and let $\mathcal{S}$ denote the corresponding \intersectionspace. The goal of the consistency algorithm is to determine whether two \intersectionspaces $\mathcal{S}_1$ and $\mathcal{S}_2$ are associated with the same activation boundary. 

We first recall the existing consistency algorithm. 
This algorithm takes two \intersectionspaces $\mathcal{S}_1$ and $\mathcal{S}_2$. 
If they are associated with the same activation boundary at depth $\ell$ and index $k$, the sum of their subspaces, $f_{1}^{(\ell-1)}(\mathcal{S}_1) + f_{2}^{(\ell-1)}(\mathcal{S}_2)$, is orthogonal to the weight vector of the corresponding neuron, $\bm w_k^{(\ell)}$. 
Here, $f_{i}^{(\ell-1)}$ denotes the local linear map up to depth $\ell-1$ induced by the input $\bm s_i$ at the \intersectionpoint. 
On the other hand, if they are not associated with the same activation boundary, no such orthogonal vector exists in the sum of subspaces. 
Thus, we compute the rank of $f_{1}^{(\ell-1)}(\mathcal{S}_1) + f_{2}^{(\ell-1)}(\mathcal{S}_2)$, and if a rank deficiency is observed, we identify that the two spaces are associated with the same activation boundary. 
Specifically, we count the number of neurons active in either $f_1^{(\ell-1)}(\bm s_1)$ or $f_2^{(\ell-1)}(\bm s_2)$. If the observed rank is lower than this number, the two \intersectionspaces are considered to belong to the same activation boundary. 

In cross-layer extraction, the local linear map must be constructed across multiple layers. 
Instead of using $f^{(\ell-1)}(\bm s)$, we introduce 
\[g^{(\ell)}(\bm s) = 
\begin{pmatrix}
    f_\mathcal{\bar P}^{(\ell)}(\bm s) \\ 
    f^{(\ell-1)}(\bm s)
\end{pmatrix},\]
where $f_{\bar{\mathcal{P}}}^{(\ell)}(\bm s)$ represents the output at layer $\ell$ with the coordinates in $\mathcal{P}$ removed. These outputs are prepended to the original $f^{(\ell-1)}(\bm s)$. 
If $\mathcal{S}_1$ and $\mathcal{S}_2$ are associated with the same activation boundary at depth $\ell+1$ and index $k$, the sum of subspaces, $g_1^{(\ell)}(\mathcal{S}_1) + g_2^{(\ell)}(\mathcal{S}_2)$, lies on the same activation boundary and is orthogonal to a weight-related vector $\bm w_{cross}$. 
\begin{align} \label{eq:orth-cross1}
    \bm w_{cross}
    ~\perp~
    \left( g_1^{(\ell)}(\mathcal{S}_1) + g_2^{(\ell)}(\mathcal{S}_2) \right).
\end{align}
Unlike the existing consistency algorithm, the cross-layer consistency algorithm requires more than two samples. 
To understand this behavior, we consider the corresponding local linear forward matrix, $\mathbf{F}^{(\ell-1)}_{\bm s}$. 
For sufficiently small $\bm \delta$, there is a matrix $\mathbf{M}$ satisfying
\begin{align}
    g^{(\ell)}(\bm s + \bm \delta) &= g^{(\ell)}(\bm s) + \mathbf{M} \bm \delta, & 
    \mathbf{M} &= 
    \begin{pmatrix}
        \mathbf{D}_{\bar{\mathcal{P}}} \mathbf{D}^{(\ell)} \mathbf{W}^{(\ell)} \mathbf{F}_{\bm s}^{(\ell - 1)} \\
        \mathbf{F}_{\bm s}^{(\ell - 1)}
    \end{pmatrix}, 
\end{align}
where $\mathbf{D}_{\bar{\mathcal{P}}}$ denotes a diagonal matrix whose $i$th diagonal entry is 1 if and only if $i \in \bar{\mathcal{P}}$.

Let $\mathbf{N}_i$ be a basis of the \intersectionspace $\mathcal{S}_i$. 
Then, the basis of the sum of subspaces is represented by the column-wise concatenation $\left( \mathbf{M}_1 \mathbf{N}_1, \mathbf{M}_2 \mathbf{N}_2 \right)$. The orthogonality condition in \cref{eq:orth-cross1} can be rewritten as
\begin{align}
    \bm w_{cross}^\top
    \left(\mathbf{M}_1 \mathbf{N}_1, \mathbf{M}_2 \mathbf{N}_2 \right) = 0,
\end{align}
or equivalently,
\begin{align}
    \begin{pmatrix}
        \mathbf{N}_1^\top \mathbf{M}_1^\top \\
        \mathbf{N}_2^\top \mathbf{M}_2^\top 
    \end{pmatrix}
    \bm w_{cross}
    = 0.
\end{align}
In cross-layer extraction, the ranks of $\mathbf{M}_1$ and $\mathbf{M}_2$ are typically low because the lower half of the row vectors are linearly dependent on those in the upper half. Therefore, the dimension of the kernel of $\mathbf M_i^\top$ is large, and
\begin{align}
    \dim\left( \ker(\mathbf M_1^\top) \cap \ker(\mathbf M_2^\top) \right) > 0
\end{align}
holds.
This implies that the intersection of the kernels, 
$\left( \ker(\mathbf M_1^\top) \cap \ker(\mathbf M_2^\top) \right)$,
contains basis vectors orthogonal to $\left( g_1^{(\ell)}(\mathcal{S}_1) + g_2^{(\ell)}(\mathcal{S}_2) \right)$ regardless of the specific choice of $\mathcal{S}_1$ and $\mathcal{S}_2$. 
In other words, the conventional consistency algorithm always observes rank deficiency\footnote{Note that this issue can happen in the original setting. For example, if there is a layer with very few active neurons before reaching depth $\ell$, $\mathrm{rank}(\mathbf{F}_{\bm s}^{(\ell-1)})$ is bounded by the number of active neurons in that layer. However, in practice, the likelihood of such false positives is significantly lower than in cross-layer scenarios, making conventional simple algorithms sufficient. }. 

\begin{algorithm}[tb]
    \caption{\texttt{IsUnifiedConsistent}($\mathbf{M}_1$, $\mathbf{M}_2$, \ldots, $\mathbf{M}_m$, $\mathbf N_1$, $\mathbf N_2$, \ldots, $\mathbf N_m$)} \label{alg:newConsistent}
    \algrenewcommand\algorithmicrequire{\textbf{Input:}}
    \algrenewcommand\algorithmicensure{\textbf{Output:}}
    \begin{algorithmic}[1] 
        \Require $\mathbf{M}_i$: a matrix to transform to the target layer, $\mathbf N_i$: the corresponding \intersectionspace basis.    % 前提条件
        \Ensure ($gap$, $score$, $\bm v_1$): statistics and the top generalized eigenvector. 
        \For{$i = 1$ to $m$}
            \State $\mathbf L_i \gets \mathbf{M}_i \mathbf{N}_i$
        \EndFor
        \State $\mathbf M = \sum_i \mathbf{M}_i \mathbf{M}_i^\top$
        \State $\mathbf L = \sum_i \mathbf{L}_i \mathbf{L}_i^\top$
        \State Solve a generalized eigenvalue problem, $\mathbf{M} \bm v = \lambda \mathbf{L} \bm v$
        \State $gap = \lambda_{1} / \lambda_{2}$                                                            \Comment{$\lambda_{\mathrm{max}} = \lambda_1 > \lambda_2 > \cdots$.}
        \State $score = (\bm v_{1}^\top \mathbf{L} \bm v_{1}) / (\bm v_{1}^\top \mathbf{M} \bm v_{1})$      \Comment{$\bm v_1$ is the maximum eigenvector.}
        \State \textbf{return} $gap$, $score$, $\bm v_{1}$
        % \State $\mathbf M = [\mathbf M_1, \mathbf M_2, \ldots, \mathbf M_m]$
        % \State $\mathbf L = [\mathbf L_1, \mathbf L_2, \ldots, \mathbf L_m]$
        % \State $\mathbf U_L = \ker(\mathbf L^\top)$  \Comment{$\mathbf{U}_L$ is orthogonal basis of $\ker(\mathbf{L}^\top)$.}
        % \State $\mathbf U_M = \ker(\mathbf M^\top)$  \Comment{$\mathbf{U}_M$ is orthogonal basis of $\ker(\mathbf{M}^\top)$.}
        % \If{$\mathrm{rank}(\mathbf U_L - \mathbf U_M \mathbf U_M^\top \mathbf U_L) = 0$}
        %     \State \textbf{return} False
        % \Else
        %     \State \textbf{return} True
        % \EndIf
    \end{algorithmic}
\end{algorithm}
To handle low-rank matrices, we propose a unified consistency algorithm. 
\cref{alg:newConsistent} shows the pseudocode. 
Given matrices $\mathbf{M}_i$ and basis matrices $\mathbf{N}_i$, we first compute the transformed bases $\mathbf{L}_i = \mathbf{M}_i \mathbf{N}_i$. 
Our goal is to determine whether there exists $\bm w_{cross}$ satisfying $\bm w^\top_{cross} \mathbf{L}_i = \bm 0$ and $\bm w^\top_{cross} \mathbf{M}_i \ne \bm 0$  for all $i$. 
To this end, we use 
\begin{align}
    \sum_i \lVert \bm w_{cross}^\top \mathbf{L}_i \rVert^2 &= \bm w_{cross}^\top \left( \sum_i \mathbf{L}_i \mathbf{L}_i^\top \right) \bm w_{cross} = 0, \\
    \sum_i \lVert \bm w_{cross}^\top \mathbf{M}_i \rVert^2 &= \bm w_{cross}^\top \left( \sum_i \mathbf{M}_i \mathbf{M}_i^\top \right) \bm w_{cross} > 0.
\end{align}
Let $\mathbf{L} = \sum_i \mathbf{L}_i \mathbf{L}_i^\top$ and $\mathbf{M} = \sum_i \mathbf{M}_i \mathbf{M}_i^\top$.
Then, maximizing $(\bm v^\top \mathbf{M} \bm v)/(\bm v^\top \mathbf{L} \bm v)$
leads to a generalized eigenvalue problem
$\mathbf{M} \bm v = \lambda \mathbf{L} \bm v$,
where the maximum eigenvector $\bm v_1$ achieves the maximum\footnote{We also tried the generalized eigenvalue problem of the form $\mathbf{L} \bm v = \lambda \mathbf{M} \bm v$, where $(\bm v_1^\top \mathbf{L} \bm v_1)/(\bm v_1^\top \mathbf{M} \bm v_1)$ is minimized. Unfortunately, this form did not work well because $\mathbf{L} \bm v = 0$ if $\mathbf{M} \bm v = 0$. Then, it yields many undetermined eigenvalues, and the minimum eigenvector is not always determined with the desired constraint. Therefore, we employed the generalized eigenvalue problem $\mathbf{M} \bm v = \lambda \mathbf{L}_{reg} \bm v$, where $\mathbf{L}_{reg}=\mathbf{L} + \epsilon \mathbf{I}$ is a regularized form of $\mathbf{L}$ to ensure numerical stability.}. 
Specifically, we check two values; the first value is $\emph{score}$ that is $(\bm v_1^\top \mathbf{L} \bm v_1) / (\bm v_1^\top \mathbf{M} \bm v_1)$, and the second value is $\emph{gap}$ that is $\lambda_1 / \lambda_2$, where $\lambda_1$ and $\lambda_2$ denote the maximum and the second maximum eigenvalues, respectively. 
The larger the score, the larger $\bm v^\top \mathbf{L} \bm v$ and the smaller $\bm v^\top \mathbf{M} \bm v$, implying that a vector $\bm w_{cross}$ satisfying the desired conditions is unlikely to exist. 
On the other hand, the larger the gap, the more significantly $\bm v^\top \mathbf{L} \bm v$ is smaller than $\bm v^\top \mathbf{M} \bm v$, implying that there is a high-confidence candidate for $\bm w_{cross}$.

\subsubsection{Experimental Verification.}
\begin{figure}[t]
    \centering
    \includegraphics[width=\linewidth]{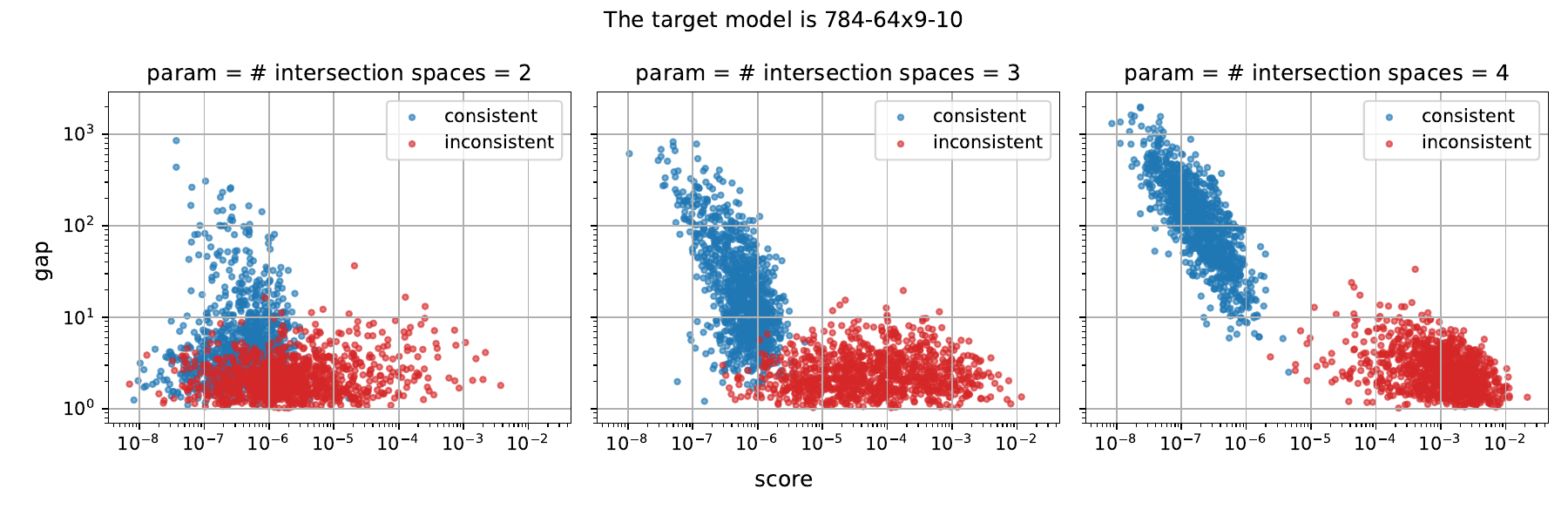}
    \includegraphics[width=\linewidth]{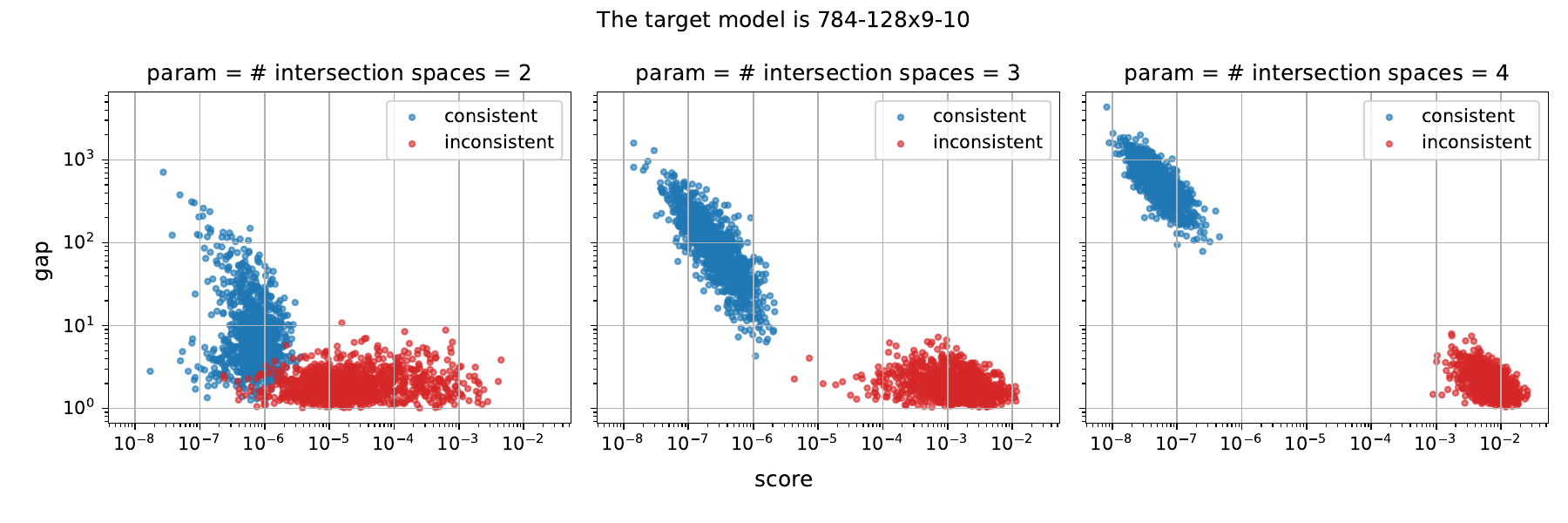}
    \includegraphics[width=\linewidth]{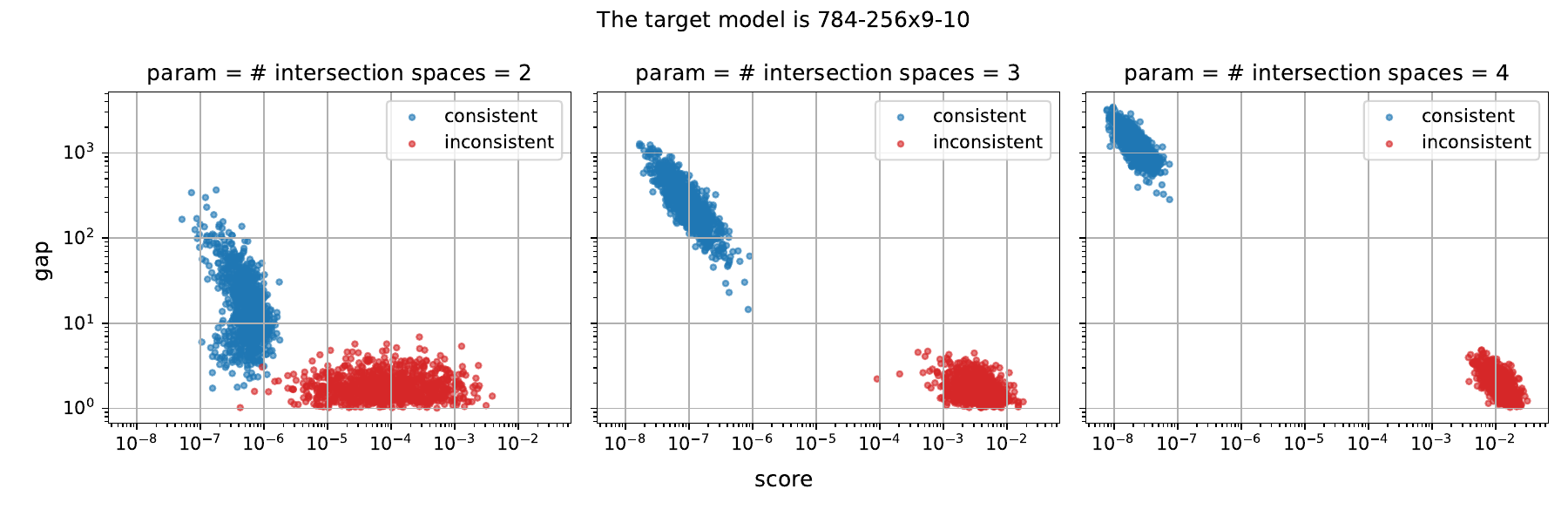}
    \caption{\texttt{IsUnifiedConsistent} for cross-layer extraction on several models, where $\epsilon = 10^{-5}$ for the regularization of $\mathbf{L}$.}
    \label{fig:unified}
\end{figure}
% \begin{table}[tb]
%     \centering
%     \caption{Accuracy of the modified consistent algorithm for cross-layer extraction. }
%     \begin{tabular}{c|c|c|c|c|c} \toprule
%         \multirow{2}{*}{model}                      & \multicolumn{4}{c}{\# \intersectionspaces}    & note\\ 
%                                                     &  2      & 3        & 4       & 5              &  \\  \midrule
%         $784 \mathrm{-} 256\times9 \mathrm{-} 10$   &29.0 \%  & 99.9 \%  & 100.0 \%  & 100.0 \%     & Target is the 4th layer and $|\mathcal{P}| = 6$. \\  \midrule
%         % $784 \mathrm{-} 128\times9 \mathrm{-} 10$   & 4.5 \%  & 71.4 \%  & 100.0 \%  & 100.0 \%     & Target is the 4th layer and $|\mathcal{P}| = 3$.    \\  \midrule
%         $784 \mathrm{-} 128\times9 \mathrm{-} 10$   & 12.1 \%  & 95.5 \%  & 100.0 \%  & 100.0 \%     & Target is the 4th layer and $|\mathcal{P}| = 2$.    \\  \midrule
%         $784 \mathrm{-} 64\times9 \mathrm{-} 10$    & 5.4\%  & 43.5 \%  & 95.2 \%  & 99.8\%        & Target is the 4th layer and $|\mathcal{P}| = 1$. \\  \bottomrule
%     \end{tabular}
%     \label{tab:prob_consistent_crosslayer}
% \end{table}
% We experimentally evaluated the unified consistency algorithm to estimate how many \intersectionspaces are required to pass the consistency check with high confidence. 
% We used trained models on the MNIST dataset. 
% We assume that the 4th layer contains persistent and dead neurons, and we performed cross-layer extraction using \intersectionpoints in the 5th layer.
We evaluated the unified consistency algorithm experimentally to estimate the number of \intersectionspaces required to pass the consistency check with high confidence, using models trained on MNIST. 
We assume the 4th layer contains persistent and dead neurons and performed cross-layer extraction from \intersectionpoints in the 5th layer.

\Cref{fig:unified} summarizes the experimental results regarding the distinguishability between consistent and inconsistent sets by \texttt{IsUnifiedConsistent}. 
We generated 1000 random consistent and inconsistent sets and applied the unified consistency algorithm. For an inconsistent set, we used a set of consistent \intersectionspaces, replacing one of them with an inconsistent \intersectionspace. 

We observe that three \intersectionspaces are sufficient to distinguish consistent and inconsistent sets for the models with 128 or 256 hidden units. 
Even for the model with 64 hidden units, four \intersectionspaces are sufficient. 
Our experiment suggests that at most one inconsistent \intersectionspace significantly changes the output of the unified consistency algorithm. Therefore, once a high-confidence consistent set is identified, we can efficiently expand it by adding a new \intersectionspace and checking for consistency. 

\paragraph{Time Complexity.}
A pairwise unified consistency algorithm can distinguish consistent and inconsistent pairs to a limited extent, but it lacks sufficient distinguishability when the number of hidden units is small. We recommend first identifying a high-confidence pair or triple, and then expand the set one by one. When we initially search over triples, the expected complexity is $O(n^3)$. 

% The complexity of the unified consistency algorithm itself is slightly higher than that of the conventional consistency algorithm. Specifically, the main difference lies in solving a generalized eigenvalue problem instead of a standard eigenvalue problem. These theoretical complexities are equivalent, while the generalized eigenvalue problem is slower by a constant factor than the eigenvalue problem for matrices of the same size. 

The unified consistency algorithm is slightly more complex than the conventional one because it solves a generalized eigenvalue problem instead of a standard one. Their theoretical complexities are equivalent, though the generalized problem is slower by a constant factor for matrices of the same size.

%\Cref{fig:unified} summarizes the experimental results regarding the accuracy of the unified consistent algorithm in the cross-layer extraction. Note that this algorithm always returns ``true'' when all \intersectionspaces are consistent. Therefore, the accuracy is evaluated based on the true negative rate, where inconsistent inputs are truly detected as inconsistent. To generate inconsistent inputs, we prepared a set of consistent \intersectionspaces and replaced one with an inconsistent one. We repeatedly applied the unified consistent algorithm and counted the number of true negatives over 1000 random samples. 

%Once a high-confidence consistent set is identified, we can efficiently expand it by adding new \intersectionspace and checking for consistency. Although fewer hidden units require more \intersectionspaces to achieve reliable consistency checks, our experiments suggest that four \intersectionspaces are sufficient in many cases. Thus, the complexity is at most $\binom{n}{4}$, where $n$ denotes the total number of \intersectionspaces. 

\subsection{Signature/Span Recovery}
Given a set of consistent \intersectionspaces, we next recover the signature, and the same \Cref{alg:newConsistent} can be used for this purpose. 
\Cref{alg:newConsistent} returns the maximum eigenvector that can be a candidate of $\bm w_{cross}$. 
When we use a sufficient number of \intersectionspaces, the rank of the matrix $\mathbf L$ is one less than the rank of the matrix $\mathbf M$. Thus, the maximum eigenvector is the only vector satisfying $\bm v^\top \mathbf L = \bm 0$ and $\bm v^\top \mathbf M \ne \bm 0$, that is $\bm w_{cross}$. 

\subsubsection{Single Persistent Neuron, $|\mathcal{P}| = 1$, and Signature Recovery.}
We discuss the simplest case, i.e., only one persistent neuron in the $\ell$th layer. Then, the maximum eigenvector is 
\begin{align}
    \bm w_{cross} = \begin{pmatrix}
        \bm w_{k,\{1,\ldots,p-1,p+1,\ldots,d_{\ell-1}\}}^{(\ell+1)} \\
        w_{k,p}^{(\ell+1)} \bm w_p^{(\ell)}
    \end{pmatrix},
\end{align}
which directly contains the signature of the persistent neuron, $\bm w_p^{(\ell)}$. 

We can successfully recover the signature, but we cannot recover the corresponding sign and bias. 
More precisely, we do not need to recover them because they do not affect the model's output as long as the neuron remains persistent.
Therefore, persistent neurons imply that ReLU is skipped. In other words, the recovered model does not have the ReLU function in persistent neurons. The sign and bias are linearly expanded into the next layer, and we can extract them together with the next-layer extraction.

\subsubsection{Multiple Persistent Neurons and Span Recovery.}
When multiple persistent neurons exist, an extra analysis is required. The maximum eigenvector contains
\begin{align}
    \sum_{j \in \mathcal{P}} w_{k,j}^{(\ell+1)} \bm w_j^{(\ell)},
\end{align}
that is a linear combination of the weights of persistent neurons. 
We repeat the cross-layer extraction against at least $|\mathcal{P}|$ target neurons in the $(\ell+1)$th layer and obtain $|\mathcal{P}|$ bases, $\tilde{\bm w}_{1}^{(\ell)}, \tilde{\bm w}_{2}^{(\ell)}, \ldots, \tilde{\bm w}_{|\mathcal{P}|}^{(\ell)}$. The span of these obtained bases is the same as the span of vectors $\bm w_j^{(\ell)}$ for $j \in \mathcal{P}$. 

We cannot recover the original basis because we cannot know $w_{k,j}^{(\ell+1)}$ as long as these neurons are persistent. 
Therefore, we use the basis $\tilde{\bm w}_{1}^{(\ell)}, \tilde{\bm w}_{2}^{(\ell)}, \ldots, \tilde{\bm w}_{|\mathcal{P}|}^{(\ell)}$ as the ``weights'' of the corresponding persistent neurons. The ReLU of these neurons must be removed in the recovered model because they were originally persistent and never suppressed by the ReLU. 
Similar to the case of $|\mathcal{P}|=1$, we do not need to consider sign and bias because there are no ReLU functions.

\subsubsection{Accuracy of Model Extraction.}
Model extraction via cross-layer extraction can, in principle, recover the target model exactly, as long as neurons classified as persistent or dead indeed remain in those respective states. If a neuron classified as persistent were to become inactive, or a neuron classified as dead were to become active, the extracted model would no longer be guaranteed to produce identical outputs to the original model.
On the other hand, the neurons we classify as persistent or dead are those whose activation states never switched between active and inactive, even after an exponential number of queries. Therefore, the extracted model reproduces the outputs of the original model with overwhelming probability.

\subsubsection{Experimental Verification.}
\begin{figure}[t]
    \centering
    \begin{subfigure}{0.32\textwidth}
        \centering
        \includegraphics[width=\linewidth]{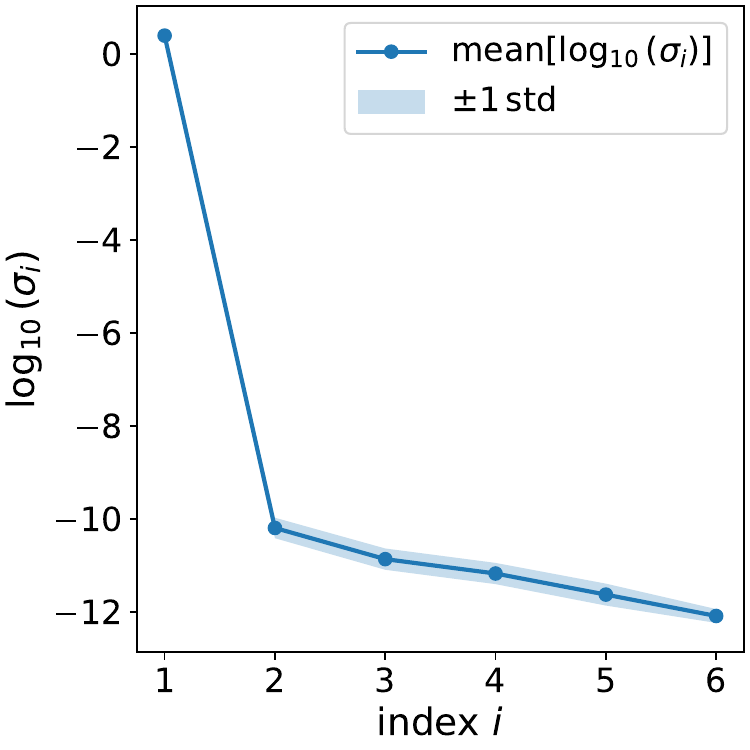}
        \caption{784-64$\times$9-10}
        \label{fig:span_recovery64}
    \end{subfigure}
    \hfill
    \begin{subfigure}{0.32\textwidth}
        \centering
        \includegraphics[width=\linewidth]{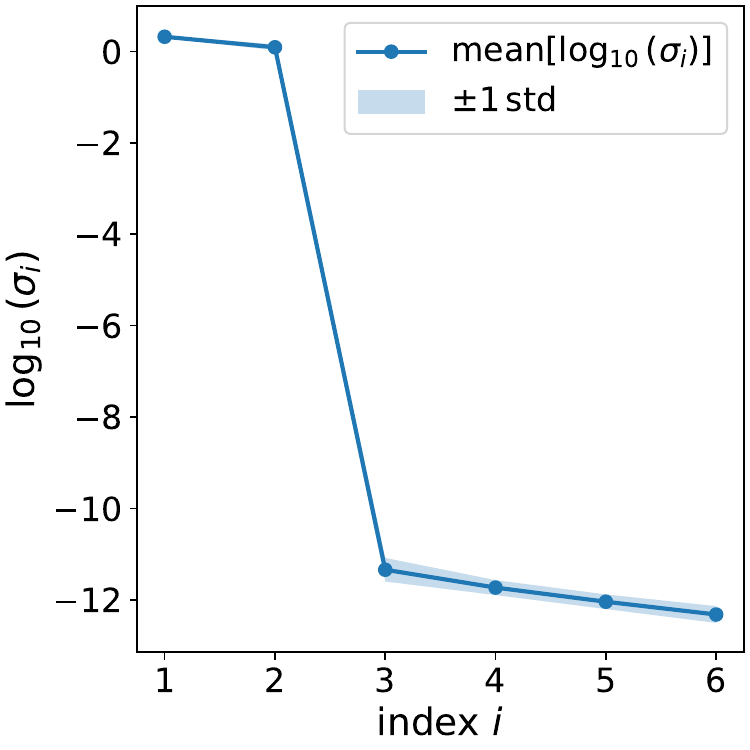}
        \caption{784-128$\times$9-10}
        \label{fig:span_recovery128}
    \end{subfigure}
    \hfill
    \begin{subfigure}{0.32\textwidth}
        \centering
        \includegraphics[width=\linewidth]{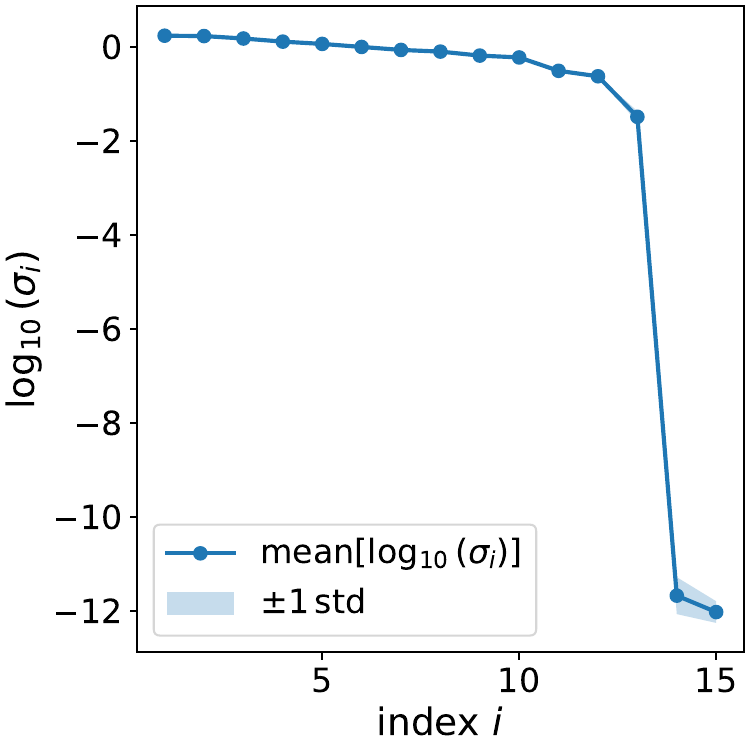}
        \caption{784-256$\times$9-10}
        \label{fig:span_recovery256}
    \end{subfigure}
    \caption{Scree plot of singular values, where $\epsilon = 10^{-13}$ for the regularization of $\mathbf{L}$.}
    \label{fig:span_recovery}
\end{figure}
To verify the feasibility of cross-layer extraction, we conducted experiments using the same model as in the unified consistency algorithm. 
We consider models in which the 4th layer contains 1, 2, and 13 persistent neurons for 64, 128, and 256 hidden units, respectively, and attempt to recover them via cross-layer extraction from the \intersectionspaces in the 5th layer.
For the models with 64 and 128 hidden units, we collected 6 vectors via cross-layer extraction, while for the model with 256 hidden units, we collected 15 vectors.
For each hidden-unit configuration, we repeated the experiment 30 times. \Cref{fig:span_recovery} shows the scree plots of the corresponding singular values.

From these experiments, we observed a clear rank deficiency, and the observed rank corresponds to the number of persistent neurons. 
We also evaluated the principal angles between the space spanned by the recovered vectors and that spanned by the true (persistent) neuron weights. The largest principal angle was approximately $10^{-8}$, which indicates successful recovery of the original span. 
%Finally, we recovered the signature and bias for the 5th layer using the recovered span. As a result, we confirmed that each neuron's output is identical to the original output after normalization. 

\subsection{Discussion}
% \subsubsection{Avoiding Local Persistent Neurons.}
% When we use \intersectionspaces, $\mathcal{S}_i$, each element of $f^{(\ell-1)}(\mathcal{S}_i)$ should exhibit both active and inactive values, unless it corresponds to a persistent or dead neuron. If a neuron is locally dead, its associated coefficient cannot be recovered. 
% If it is locally persistent, its weight may leak into the linear combination of persistent neurons. 
% Once it occurs, removing its influence becomes difficult because its weight is not necessarily orthogonal to those of the persistent neurons. 

% Therefore, it is necessary to obtain $|\mathcal{P}|$ neurons that do not contain any local persistent neurons. This is expected to be feasible, as the corresponding signature and signs were successfully extracted during the attack on the preceding layer. In other words, they are neither persistent nor dead. 
% Nevertheless, if we cannot collect a sufficient number of neurons free from local persistence, the best possible strategy is to treat them as part of the persistent neurons. The resulting recovered model still produces correct outputs, provided that the input satisfies the conditions of local persistence. 
\subsubsection{Avoiding Local Persistent Neurons.}
When using \intersectionspaces $\mathcal{S}_i$, each coordinate of $f^{(\ell-1)}(\mathcal{S}_i)$ should attain both active and inactive values, unless it corresponds to a persistent or dead neuron. If a neuron is locally dead, its coefficient cannot be recovered. If it is locally persistent, its weight may be absorbed into the linear combination of persistent neurons. Once this happens, removing its influence is difficult because its weight is not necessarily orthogonal to the persistent neurons.

Therefore, we need $|\mathcal{P}|$ target neurons that are free of local persistence. This is expected to be feasible, since their signatures and signs were successfully extracted in the preceding-layer attack, implying that they are neither persistent nor dead. If we still cannot collect enough such neurons, the best strategy is to treat them as persistent; the recovered model then remains correct as long as local persistence holds for the input.

% \subsubsection{Recovering Future Neurons.}
% The cross-layer extraction recovers the signature or span of persistent neurons. Once these are recovered and appropriate weights are assigned, followed by the removal of ReLU, the attack is considered complete up to the $\ell$th layer. From the next layer, we go back to the original attack procedure. 
% As long as there is no contradiction in the classification of persistent or dead neurons, we do not have any problem in the attack for the next layer. In an extreme (though unlikely) case where all neurons in the next layer can equally exhibit both active and inactive states, we can fully come back to the original model extraction procedure.  
% In the more common scenario, where persistent or dead neurons are present in the next layer, we first recover signatures of all neurons except persistent or dead neurons. We next recover their sign. We finally apply cross-layer extraction to recover the signature/span for persistent neurons. 

% It is important to emphasize that the recovered model is not strictly identical to the original model. During the model extraction, the attacker uses only the case, where neurons exhibit persistent behavior. In other words, if an input has contradictory activation patterns in persistent/dead neurons, the recovered model may produce different outputs from the original model.  

\subsubsection{Recovering Future Neurons.}
Cross-layer extraction recovers the signature or span of persistent neurons. Once these are recovered, we assign appropriate weights and remove the ReLU, completing the attack up to the $\ell$th layer. From the next layer onward, we return to the original extraction procedure.

As long as the classification of persistent/dead neurons is consistent, the subsequent layers pose no additional difficulty. If no persistent or dead neurons appear in the next layer, we fully revert to the original procedure. More commonly, when persistent or dead neurons are present, we (i) recover signatures of all recoverable neurons, (ii) recover their signs, and (iii) apply cross-layer extraction to obtain the signature/span of persistent neurons.

The recovered model is not strictly identical to the original model. The attacker exploits only inputs on which persistent/dead neurons exhibit the assumed behavior; thus, if an input induces contradictory activation patterns, the recovered model may produce different outputs.

\subsubsection{On the Soft-Label Setting.}
Our main focus is the hard-label setting, but attacks on the soft-label setting have also been studied~\cite{DBLP:conf/crypto/CarliniJM20,DBLP:conf/eurocrypt/CanalesMartinezCHRSS24,DBLP:conf/asiacrypt/ChenDMSWYW25,DBLP:conf/eurocrypt/LiuSELBP26}. We believe that cross-layer extraction or its extension is also helpful in the soft-label setting.
In the soft-label setting, one can compute second-order derivatives of the output logits, which directly reveal activation boundaries (without going through the decision boundary). However, similar to the hard-label setting, activation boundaries associated with persistent neurons are inaccessible from the input, which causes inevitable errors in the recovered forward local linear map.
Similar to the hard-label setting, the soft-label attacker also must peel off already recovered layer to attack the subsequent layer. Here, since the only available recovered map contains errors, we face the same issue.
Since the soft-label attacker is strictly more powerful than the hard-label attacker, it is reasonable to expect that techniques analogous to cross-layer extraction could also be applied to address this issue in the soft-label setting. 
We leave the development of a methodology specialized for the soft-label setting as an open question.

% Since the soft-label attacker is strictly more powerful than the hard-label attacker, the cross-layer extraction can solve this issue in the soft-label setting as well. On the other hand, there may be a more efficient solution specific to the soft-label setting, but this remains an open question.

% Our main focus is the hard-label setting, but attacks on the soft-label setting have also been studied~\cite{DBLP:conf/crypto/CarliniJM20,DBLP:conf/eurocrypt/CanalesMartinezCHRSS24,DBLP:journals/corr/abs-2506-17047,DBLP:conf/asiacrypt/ChenDMSWYW25}. 
% In the soft-label setting, one can compute second-order derivatives of the output logits, which directly reveal activation boundaries (without going through the decision boundary). However, similar to the hard-label setting, activation boundaries associated with persistent neurons are inaccessible from the input, which causes inevitable errors in the recovered forward local linear map. 
% To recover full signatures in subsequent layers, we transform them using the recovered (erroneous) map and merge the transformed signatures to complement missing coefficients. Here, we face the same issue as in the hard-label setting. 
% Since the soft-label attacker is strictly more powerful than the hard-label attacker, this issue can be addressed by adopting an approach analogous to cross-layer extraction. On the other hand, there may be a more efficient solution specific to the soft-label setting, but this remains an open question. 

\section{Conclusion}
In this paper, we revisited the hard-label cryptanalytic model extraction proposed at Eurocrypt 2025 \cite{DBLP:conf/eurocrypt/CarliniCHRS25}. 
Carlini et al. claimed that the attack runs in polynomial time under the assumption that all neurons switch between active and inactive states with equal probability. 
However, we showed that this assumption is unlikely to hold in practice. In particular, the existence of persistent (always active) neurons introduces non-negligible and unavoidable noise, which can render the attack infeasible. 
To address this issue, we proposed \emph{cross-layer extraction}, which recovers a model that behaves identically to the original as long as persistent and dead neurons remain active and inactive, respectively. 
%If there are $n$ $\varepsilon$-persistent neurons and $m$ $\varepsilon$-dead neurons, the recovered model is correct with probability $(1-\varepsilon)^{n+m}$. 
If there are $n$ $\varepsilon$-persistent neurons and $m$ $\varepsilon$-dead neurons, then by the union bound, the recovered model is correct with probability at least $1-(n+m) \varepsilon$.

Finally, we emphasize that both our attack and the original attack are entirely theoretical, assuming that \intersectionpoints can be extracted without noise whenever they exist. In practice, however, unavoidable numerical errors, such as machine epsilon, tend to accumulate as deeper layers are analyzed. Therefore, from an engineering perspective, it remains an open problem whether the attacker can reliably extract these models under realistic noise conditions.

\subsubsection*{\ackname}

This work was supported by JST K Program Grant Number JPMJKP24C1, Japan.

\bibliographystyle{splncs04}
\bibliography{main.bib}

\iflncs
\else
\clearpage
\appendix

\section{Additional Experimental Results}

\subsection{Switching Probability and Proportion of Intersection Points}
\cref{fig:all_layer_MNIST_0,fig:all_layer_MNIST_30,fig:all_layer_FMNIST_30} show the relationship between the switching probability of each neuron and the number of discovered \intersectionpoints across all layers. Each point corresponds to a neuron, where the vertical axis indicates the number of discovered \intersectionpoints normalized by the total number of explored points, and the horizontal axis represents the switching probability of each neuron. Under all experimental conditions, a strong correlation between the two quantities can be observed.

\begin{figure}[t]
    \centering
    \includegraphics[width=\linewidth]{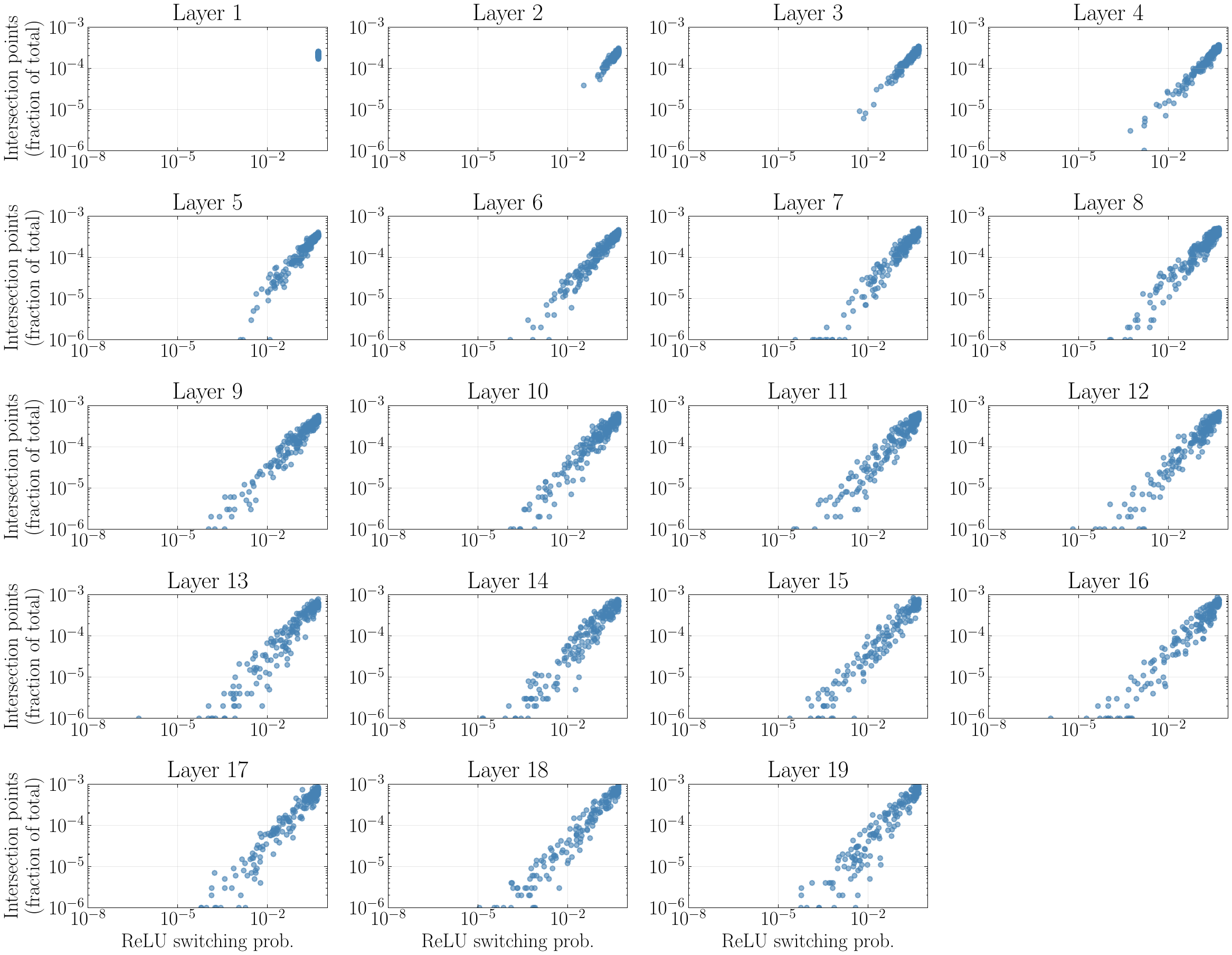}
    \caption{Switching probability versus the proportion of discovered intersection points in all layers of an untrained MLP.}
    \label{fig:all_layer_MNIST_0}
\end{figure}

\begin{figure}[t]
    \centering
    \includegraphics[width=\linewidth]{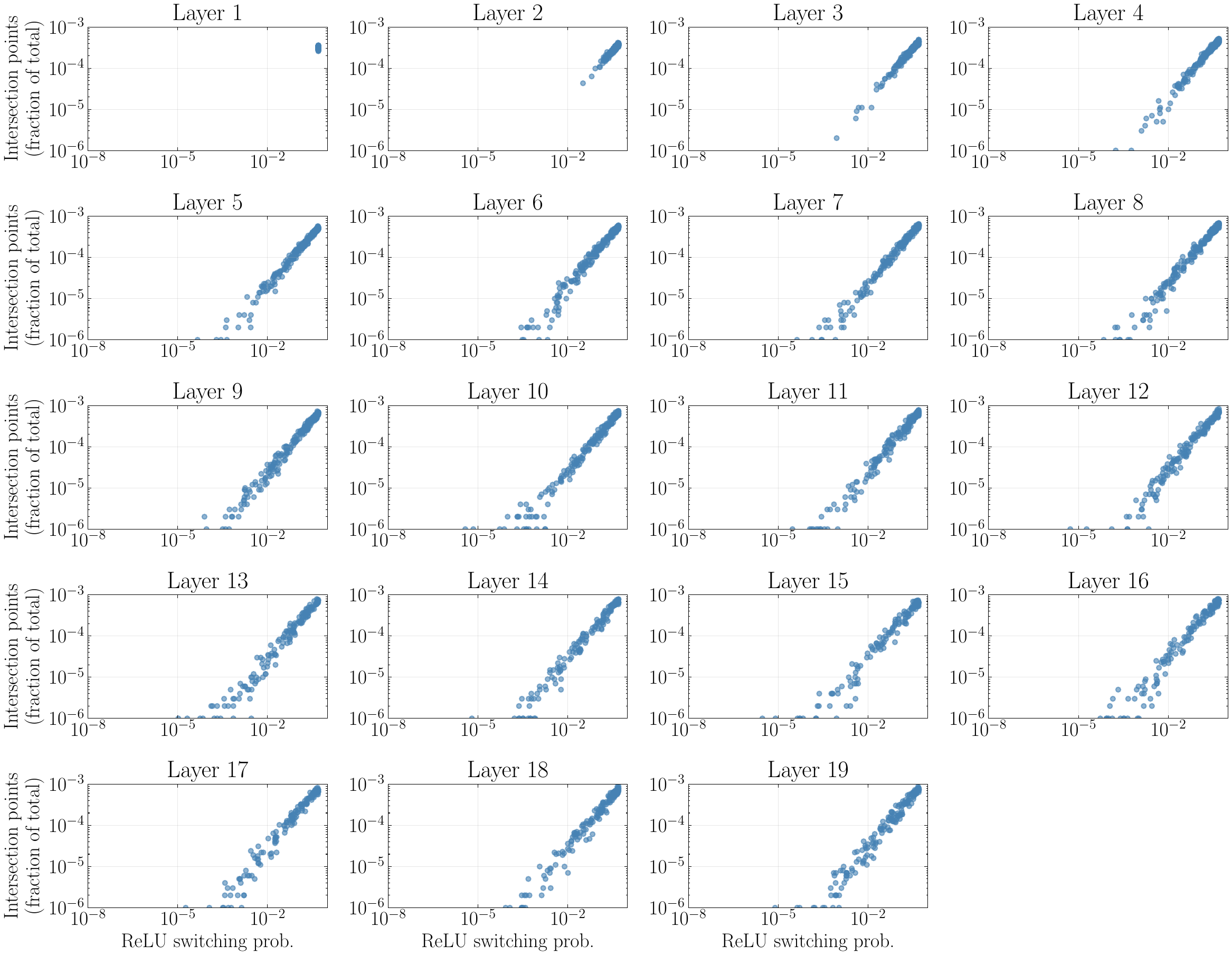}
    \caption{Switching probability versus the proportion of discovered intersection points in all layers of an MLP trained on MNIST.}
    \label{fig:all_layer_MNIST_30}
\end{figure}

\begin{figure}[t]
    \centering
    \includegraphics[width=\linewidth]{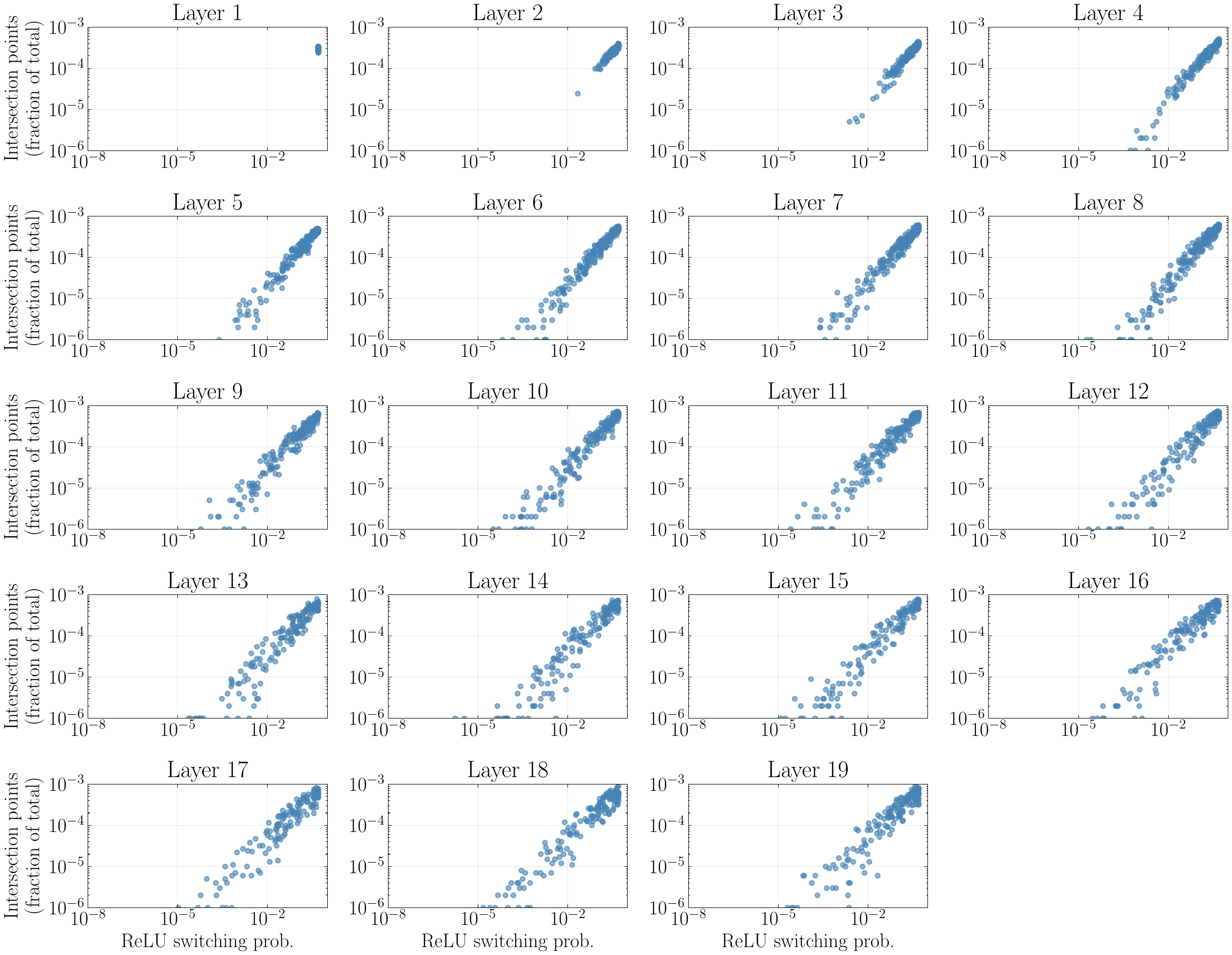}
    \caption{Switching probability versus the proportion of discovered intersection points in all layers of an MLP trained on FMNIST.}
    \label{fig:all_layer_FMNIST_30}
\end{figure}

%\section{Source Code}
%The source code used to verify the correctness of the unified consistency algorithm and span recovery in cross-layer extraction is provided in the supplementary material. Please refer to the included README file for instructions.

\fi
\end{document}